\documentclass{article}

\usepackage[final,nonatbib]{neurips_data_2024}
\usepackage[numbers]{natbib}

\usepackage[utf8]{inputenc} %
\usepackage[T1]{fontenc}    %
\usepackage{hyperref}       %
\usepackage{url}            %
\usepackage{booktabs}       %
\usepackage{amsfonts}       %
\usepackage{nicefrac}       %
\usepackage{microtype}      %
\usepackage{xcolor}         %
\usepackage{graphicx}
\usepackage{tabularx}
\usepackage{multirow}
\usepackage{subcaption}
\usepackage{wrapfig}

\newcommand{\rebuttal}[1]{{#1}}

\title{ClevrSkills: Compositional Language and Visual Reasoning in Robotics}

\author{%
  Sanjay Haresh \\
  Qualcomm AI Research\thanks{Qualcomm AI Research is an initiative of Qualcomm Technologies, Inc.} \\
  \texttt{sanjayh@qti.qualcomm.com} \\
  \And
  Daniel Dijkman \\
  Qualcomm AI Research \\
  \texttt{ddijkman@qti.qualcomm.com} \\
  \AND
  Apratim Bhattacharyya \\
  Qualcomm AI Research \\
  \texttt{aprabhat@qti.qualcomm.com} \\
  \And
  Roland Memisevic \\
  Qualcomm AI Research \\
  \texttt{rmemisevic@qti.qualcomm.com} \\
}

\begin{document}

\maketitle

\begin{abstract}
Robotics tasks are highly compositional by nature. For example, to perform a high-level task like cleaning the table a robot must employ low-level capabilities of moving the effectors to the objects on the table, pick them up and then move them off the table one-by-one, while re-evaluating the consequently dynamic scenario in the process. 
Given that large vision language models (VLMs) have shown progress on many tasks that require high level, human-like reasoning, we ask the question: if the models are taught the requisite low-level capabilities, can they compose them in novel ways to achieve interesting high-level tasks like cleaning the table without having to be explicitly taught so?
To this end, we present ClevrSkills - a benchmark suite for compositional reasoning in robotics. 
ClevrSkills is an environment suite developed on top of the ManiSkill2~\cite{gu2023maniskill2} simulator and an accompanying dataset. 
The dataset contains trajectories generated on a range of robotics tasks with language and visual annotations as well as multi-modal prompts as task specification. 
The suite includes a curriculum of tasks with three levels of compositional understanding, starting with simple tasks requiring basic motor skills. 
We benchmark multiple different VLM baselines on ClevrSkills and show that even after being pre-trained on large numbers of tasks, these models fail on compositional reasoning in robotics tasks.
\end{abstract}

\section{Introduction}

Compositional generalization is a hallmark feature of human intelligence. Unlike any other animals, humans can receive instructions in natural language and successfully perform previously unseen tasks with minimal to no task-specific learning or adaptation. Modeling this capability has been a long-standing aspiration in AI, dating back at least to Winograd’s influential SHRDLU system \cite{winograd1972understanding} developed more than half a century ago. 
The architectural underpinnings that enable these capabilities in humans have remained an inspiration as well as puzzle until this day 
\cite{kahneman2011thinking,riveland2024natural}.

A potential steppingstone towards replicating this ability in AI systems is the recent progress in language modeling, based on large models pre-trained using next-token-prediction. These models have shown encouraging compositional reasoning behaviors in response to language-based prompts – an ability that was confined initially to text-based tasks, but that since has been extended to multi-modal, and most recently also to robotics tasks. %

Compositional reasoning based on language has evolved hand-in-hand with the introduction of benchmark tasks and challenges. For language-based reasoning tasks, these include, for example, the bAbI AI challenge \cite{weston2015aicomplete}, GSM8k \cite{cobbe2021gsm8k}, and many others. 
Multi-modal tasks include the popular CLEVR challenge \cite{johnson2017clevr} and its descendants (eg., \cite{Yi*2020CLEVRER:,bahdanau2020closure}), and various intuitive physics datasets (eg., \cite{pmlr-v48-lerer16,phys101,goyal2017something}). Common to these challenges is that they require a model to reason about a scene or situation. 

Despite their reliance on some degree of ``common sense'', these existing challenges do not require 
any type of actions, behaviors or planning. As such, they are confined to evaluating compositionality in a purely abstract setting, even in the case where the input data is multi-modal. 
In this work, we propose an environment and corresponding suite of tasks, which instead allow us to study compositional generalization in a highly controlled, but complex robotics context. 
Our benchmark is based on dexterous manipulation tasks, such as pick, place, throw, touch and push within the ManiSkill2 simulation environment~\cite{gu2023maniskill2}, and it evaluates the ability to generalize to complex tasks based on these low-level capabilities.  

Our benchmark allows us to assess a model's capability to perform compositional generalization 
with respect to the creation and execution of step-by-step execution plans. 
However, unlike existing benchmarks, such as Vima \cite{jiang2023vima}, our benchmark includes not just the higher-level planning but also the low-level execution layers for a wide variety of end-to-end robotics tasks. This allows us to assess not just a model's ability to perform abstract planning in isolation but a model's ability to plan-and-execute within a closed loop.

Our contributions in detail are as follows:
\begin{itemize}
    \item We introduce the ClevrSkills\footnote{Data and code are available at \url{https://www.qualcomm.com/developer/software/clevrskills-dataset} and \url{https://github.com/Qualcomm-AI-research/ClevrSkills}} environment suite, consisting of 33 different tasks spread across 3 different levels which can be used to benchmark compositional reasoning in robotics models.
    \item We introduce an accompanying dataset of 330k ground truth trajectories generated by scripted oracle policies which use motion planning to achieve the tasks that can be used for imitation learning. The dataset also contains many types of annotation, including language, action classes, bounding boxes for objects, visibility annotations, key-steps, rewards (for offline RL), camera parameters and more.
    \item We benchmark SOTA open-source vision language models and show that they tend to fail on tasks requiring compositional understanding.
\end{itemize}

\section{Related Work}

\subsection{Vision language models for robotics}
Large vision language models, including LLaVA~\cite{liu2023llava} and others, have shown strong zero-shot and few-shot generalization %
across a wide range of tasks. 
Unsurprisingly, there has been an increasing effort to get similar results in robotics. For example, CLIPort~\cite{shridhar2022cliport}, Perceiver-Actor~\cite{shridhar2023perceiver}, or RT-1~\cite{brohan2022rt} introduce large transformer models for a range of robotics tasks. RT-2~\cite{brohan2023rt} takes this further by co-finetuning a language model on both internet scale text data and large scale robotics data. GATO~\cite{reed2022generalist} and JAT~\cite{gallouedec2024jack} similarly train a transformer based model that can work across many different tasks and modalities. Octo~\cite{team2024octo} is another recent work that proposes a transformer based generalist policy that can be finetuned on downstream tasks. RoboFlamingo~\cite{li2023vision} is based on finetuning off-the shelf VLMs on robotics data to show that they are effective at imitation learning. 
Furthermore, above-mentioned Vima~\cite{jiang2023vima} benchmarks the capability of these models to generalize in highly controlled robotics tasks. Our work is  similar in spirit, but goes beyond it in that it tests the ability of these models to generalize to not only new objects/textures/scenes as in Vima but also to totally new tasks given a base set of skills that are sufficient to complete the higher levels tasks.

\subsection{Simulators/Benchmarks}

There has been a host of simulators introduced in recent years studying various aspects of robot learning. 
This includes, for example, iGibson~\cite{li2021igibson}, Habitat2.0~\cite{szot2021habitat}, Ai2THOR~\cite{kolve2017ai2}, Behavior1k~\cite{li2023behavior}, which all support indoor environments with tasks ranging from visual goal navigation, to mobile manipulation, to re-arrangement, etc. ManiSkill2~\cite{gu2023maniskill2}, ManiPose~\cite{yu2024manipose}, DexArt~\cite{bao2023dexart} all introduce different manipulation benchmarks. Meta-world~\cite{yu2020meta} describes a benchmark for meta-reinforcement learning in table-top environments. CALVIN~\cite{mees2022calvin} and Arnold~\cite{gong2023arnold} present language-conditioned long-horizon table-top tasks that require skill chaining for success. %

Our benchmark is similar in spirit to the Vima benchmark~\cite{jiang2023vima}, with several important differences. Vima is, to the best of our knowledge, the first robotics benchmark supporting multi-modal task specifications and a controlled probing of agent capabilities. However, the benchmark is limited in that the action space is composed of object poses instead of pose deltas of the robot end-effector and consequently there is no support for assessing compositionality of the agent given a base skill set. 
We overcome this limitation in ClevrSkills to enable benchmarking of compositional reasoning. 

Most of the existing simulators and benchmarks either focus on just the manipulation skills (e.g. ManiSkill2~\cite{gu2023maniskill2}), or they abstract the manipulation skills away using oracle policies and only test a model's ability to use the oracle policies to achieve complex tasks (e.g. Vima~\cite{jiang2023vima}). The goal of ClevrSkills is to combine the best of both worlds and to enable training and benchmarking an agent's ability to acquire manipulation skills and to compose them in novel ways to solve higher-level tasks.

\section{ClevrSkills Environment Suite}

\begin{figure}
  \centering
  \includegraphics[width=0.99\linewidth]{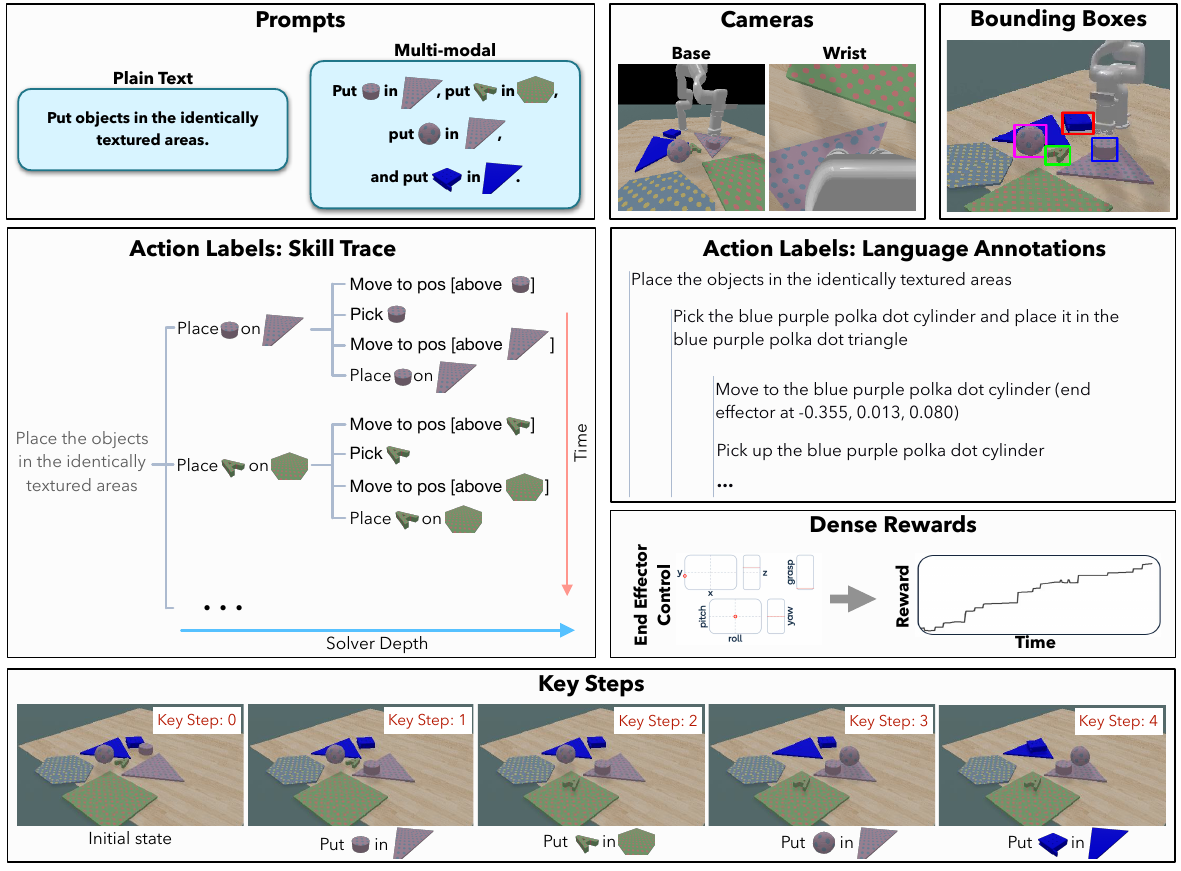}
  \caption{The ClevrSkills environment suite includes support for multi-modal prompts as task specification, multi-camera RGB observations, dense hierarchical action labels, action demonstrations in end-effector space and support for RL with dense rewards for all the tasks.}
  \label{fig:features}
  \vspace{-0.2cm}
\end{figure}

ClevrSkills is built within the ManiSkill2 simulator, which allows for realistic physics and graphics. We use a simulated model of the UFACTORY xArm 6 robot with vacuum gripper as our default robot for the environments, with Franka Emika Panda also being available. 
\begin{figure}
\centering
\includegraphics[width=0.9\linewidth]{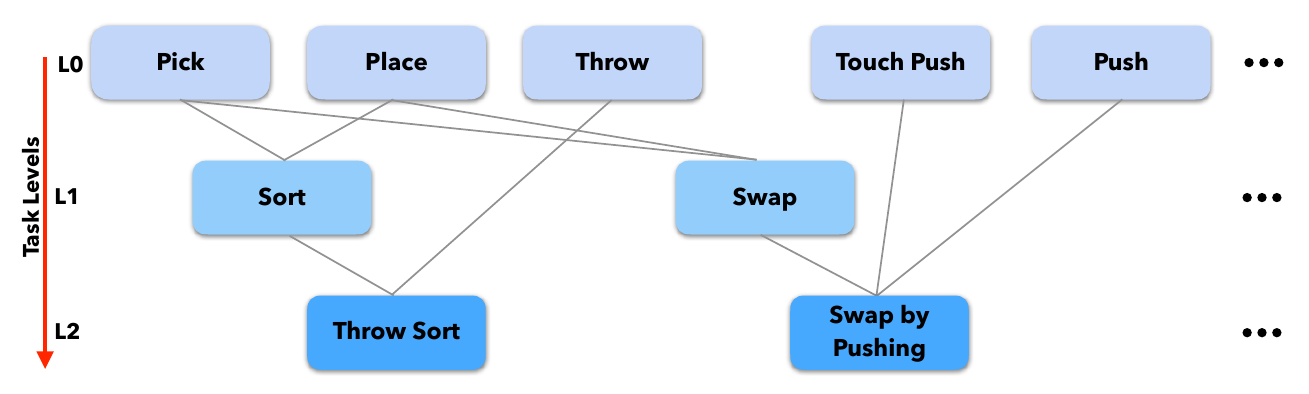}
\caption{Example task compositions in ClevrSkills. Higher level tasks in ClevrSkills are built on skills acquired from lower level tasks (L0 $\rightarrow$ L1 $\rightarrow$ L2).}
\label{fig:task_comp}
\end{figure}

We add support for multi-modal prompts for task specification, add language annotations for the actions of the robot policy, extend the objects and texture databases and add a multitude of tasks that require increasingly higher levels of compositional understanding. A snapshot of the ``Sort'' task along with the observation and action space, task specification and action labels is shown as an example in Figure~\ref{fig:features}.
A comparison to other simulator and datasets can be seen in Table~\ref{tab:stats}. 

\subsection{Task Suite}

We develop 33 different tasks carefully designed to test compositional generalization of robotics models in a highly controlled setting. 
Our task set includes simple manipulation tasks (e.g. \textit{moving from A to B}, \textit{pick}, \textit{place}, \textit{push},  \textit{tracing path}) which allow the model to learn basic manipulation/motor skills, intermediate tasks (e.g. \textit{sorting objects by texture}, \textit{stacking}), which test model's ability to compose the manipulation skills learned from simple tasks, and finally complex tasks (e.g. \textit{stacking and toppling structures}, \textit{sorting by throwing}, \textit{balancing scales with weights}), which require higher level compositional reasoning. For example, the model needs to make use of the \textit{throwing} skill learned from the first set of tasks and the \textit{sorting by texture} capability learned from the second set and then compose these two skills to successfully solve the \textit{sorting by throwing} task. This design of compositional tasks is shown in Figure~\ref{fig:task_comp}. We provide specific details of these levels in Section~\ref{sec:bench}.

\subsection{Predicates}

The reward and success criteria for the individual tasks are specified using \textit{predicates}. There are two main types of predicates: physical and logical.

Physical predicates specify the target state of the robot and/or the objects in the scene, and how the agent achieves these states. \textit{EEAtPos} and \textit{EEAtPose} require the end-effector to be at a specified position or pose (within some specified tolerance). \textit{AtPos} and \textit{AtPose} require an object to be at a specified position or pose. \textit{OnTop} and \textit{Inside} require an object to be on top or inside another object. \textit{Touch} requires the agent to touch or push an object. \textit{Hit} requires the agent to drop or topple an object onto another object.
\textit{ToppleStructure} requires a collection of objects to be on the ground. 

Logical predicates can be used to combine physical predicates to specify more complex tasks. The logical predicates are \textit{Set} (all sub-predicates must be completed in any order), \textit{Sequence} (sub-predicates must be completed in order), and \textit{Once} (the sub-predicate must be completed once). 

The dense reward of physical predicates is designed to allow RL agents to learn tasks. See the plot of the (instantaneous) dense reward shown in Figure~\ref{fig:features} for an example. Logical predicates aggregate the rewards of their sub-predicates as appropriate. Note that the decomposition of tasks into predicates allows ClevrSkills to be easily extendable as new tasks can be easily specified as compositions of these predicates.

\subsection{Oracle policies}

We develop oracle policies for all the tasks in our environment suite. These policies are called \textit{solvers} as they are designed to solve the predicates that define the tasks. The top-level solver algorithm performs a greedy search for the next predicate to solve, and instantiates a solver policy for the same. 

The mapping from predicate to a specific solver is scripted manually. The available solvers are \textit{Pick}, \textit{Place}, \textit{Move}, \textit{Trace}, \textit{Touch}, \textit{Push}, \textit{Hit} (throw object towards other object), \textit{ToppleStructure}, \textit{BalanceScale} (place objects on a scale to balance it). The \textit{Move} solver internally uses the MPLib~\cite{mplib} motion planning library, which is a Python wrapper around the implementation of RRT algorithm \cite{rrt} found in OMPL \cite{ompl}.

Higher-level solvers internally use other solvers. For example, the \textit{PickMovePlace} solver internally uses \textit{Move} to get the end-effector close to a position where it can pick the object, \textit{Pick} to pick up the object, \textit{Move} to carry the object close to the target, and \textit{Place} to place the object.  Solvers will typically let their sub-solvers take actions in the environment until the sub-solver reports that it has completed the action or has failed. 

Because the solver policies are stateless, they can be combined with other policies. E.g., one can start collecting oracle solver trajectories from states that were reached by an RL agent (e.g., to perform fine-tuning using an approach like DAgger~\cite{ross2011reduction})

\subsection{Observations and action space}
Since we extend ManiSkill2, we inherit its flexible observation and action space. However, for the purposes of this benchmark, we constrain the observation space to be RGB images from two cameras: an end-effector mounted camera which gives the robot's ``first-person'' view of the scene, and a base camera which provides a ``third-person'' perspective. 
The action space is restricted to the \textit{delta end-effector pose} controller from ManiSkill2, which provides for a 6DOF pose delta and 1D gripper scalar value as shown in Figure~\ref{fig:features}. Note that delta end-effector action demonstrations are convertible to any other controller type supported by ManiSkill2.

\subsection{Annotations} 
The success of recent large vision-language models is largely due to the availability of large amounts of paired vision and language data. 
However, such data is lacking in the case of robotics. 
Therefore, we also provide fine-grained language annotations for each step the oracle policy takes to complete any task. We provide three levels of language annotations including task or predicate level (the highest level describing the task, which can also be used as the task specification), 
sub-task level (a sub-task on a semantic level that needs to be achieved for the high level task to be completed), and step level (a language label for each  step that is being taken). The hierarchy of language annotations can be seen in Figure~\ref{fig:features} (middle).

We also provide bounding boxes and visibility labels for each object at each time-step of the generated trajectories as seen in Figure~\ref{fig:features} (top-right), as well as key-step frames corresponding to the completion of sub-tasks (Figure~\ref{fig:features} bottom section).

\subsection{Dataset} 
We generate 10k trajectories for each task using the corresponding oracle policy, resulting in a total of $\approx$330k trajectories. We split the set of objects and textures into train and test splits to test the OOD generalization of models to unseen objects and textures. Each of our tasks is used both in training and testing according to the evaluation protocol described in Section~\ref{sec:bench}. The dataset is available at \url{https://www.qualcomm.com/developer/software/clevrskills-dataset}.

\begin{table*}[t]
  \centering
  \scriptsize
  \begin{tabularx}{\linewidth}{lc@{\hspace{0.3cm}}c@{\hspace{0.3cm}}c@{\hspace{0.3cm}}c@{\hspace{0.3cm}}cc}
    \toprule
    \textbf{Dataset/Simulator} & \textbf{\#Tasks} & \textbf{Language} & \textbf{Multimodal Prompts} & \textbf{Action Granularity} & \textbf{Compositionality} & \textbf{\#Demonstrations} \\
    \midrule
    \multicolumn{7}{c}{{Real}} \\
    \midrule 
    RoboTurk~\cite{mandlekar2018roboturk}  & $3$ & $\times$ & $\times$ & Action Deltas & $\times$ & 111hrs\\
    BridgeData~\cite{ebert2021bridge} & $71$ & $\times$ & $\times$ & Action Deltas & $\times$ & 7.2k\\
    Open-X~\cite{padalkar2023open} & - & \checkmark & $\times$ & Action Deltas & $\times$ & 1M\\
    RH20T~\cite{fang2023rh20t} & - & \checkmark & $\times$ & Action Deltas & $\times$ & 100k \\
    FMB~\cite{luo2024fmb} & $7$ & $\times$ & $\times$ & Action Deltas & \checkmark & 22.5k\\
    \midrule
    \multicolumn{7}{c}{{Simulated}} \\
    \midrule
    CALVIN~\cite{mees2022calvin} & $34$\* & \checkmark & $\times$ & Action Deltas & \checkmark$\dagger$ & -- \\
    Behaviour-1K~\cite{li2023behavior} & $1000$ & $\times$ & $\times$ & Action Deltas & $\times$ & --\\
    Maniskill2~\cite{gu2023maniskill2} & $20$ & $\times$ & $\times$ & Action Deltas & $\times$ & $\approx$70k\\
    VIMA~\cite{jiang2023vima} & $17$ & \checkmark & \checkmark & Poses & $\times$ & 650k \\
    ClevrSkills (our) & $33$ & \checkmark & \checkmark & Action Deltas + Poses & \checkmark & 330k\\
    
    \bottomrule
  \end{tabularx}
  \caption{Comparison of datasets/simulators. $\dagger$ Compositionality in CALVIN mainly refers to stitching of sub-tasks to achieve long horizon tasks.}
  \label{tab:stats}
\end{table*}

\begin{figure*}[t!]
    \centering
    \begin{subfigure}[t]{0.5\textwidth}
        \centering
        \includegraphics[width=1.0\linewidth,trim={0.6cm 13.5cm 22.1cm 0.5cm},clip]{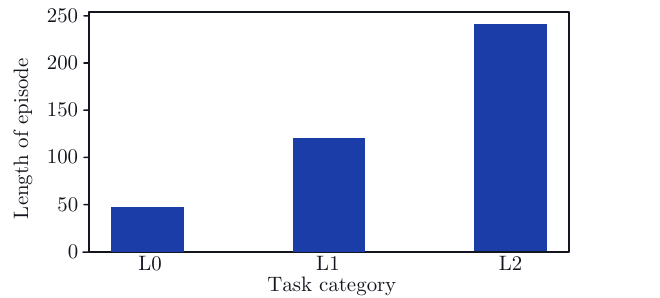}
    \end{subfigure}%
    ~ 
    \begin{subfigure}[t]{0.5\textwidth}
        \centering
        \includegraphics[width=1.0\linewidth,trim={11.3cm 13.5cm 11.4cm 0.5cm},clip]{figures/task_complexity_v2.pdf}
    \end{subfigure}
    \vspace{-0.5cm}
    \caption{\emph{Left}: The median length of an episode across task levels showing significant increase in episode length as we go from lower to higher levels of compositionality. \emph{Right}: The mean number of solvers used by the oracle to complete a task across task levels. Each solver solves for a specific sub-task, showing higher levels have increasingly compositional tasks.}
    \label{fig:task_complexity}
\end{figure*}

\section{Benchmark}
\label{sec:bench}

Our environment suite consists of 33 tasks across three levels of difficulty:
\begin{itemize}
    \item \textbf{L0: Simple Tasks.} 12 tasks that teach the agent a base set of motor skills like pick, place, throw, touch, push which can then be used to perform more complicated tasks.
    \item \textbf{L1: Intermediate Tasks.} 15 tasks that test the agent's ability to compose the skills learned from the simple tasks to perform simple compositions, such as sorting objects, stacking, swapping, rotating etc. 
    \item \textbf{L2: Complex Tasks.} 6 tasks that require long-range compositional understanding which test the models ability to compose skills learned from both the Simple and Intermediate subsets to achieve more complicated goals, such as balancing a scale with weights, sorting by throwing, swapping by pushing, etc. 
\end{itemize}
The increasing complexity of these tasks can be seen in Figure~\ref{fig:task_complexity}. A full list of tasks along with their specification and success criteria can be found in the Appendix~\ref{app:tasks}.

\textbf{Task Specification.} We follow Vima~\cite{jiang2023vima} to support multi-modal prompts (interleaved text and images) as task specification as well as with text-only prompts. 

\textbf{Evaluation.} Our main goal with this benchmark is to test the ability of vision language models to compose simple motor skills in novels ways to perform more complex tasks, both zero-shot and using fine-tuning. 
We use a three level protocol to systematically test the compositional abilities of the models. At each level, we further evaluate the models on seen and unseen attributes (objects, textures and object placements). 
The environment also provides partial rewards at each step along with binary success criteria. We report both success rate and average reward achieved for each task. 

\begin{enumerate}
    \item L0: Here, we test the model's ability to pick the base motor skills required to solve higher level tasks. All the prompts are seen at training time. 
    \item L0 -> L1: Here, we test the model's ability to compose skills from L0 tasks to achieve L1 tasks, both zero-shot and using fine-tuning. 
    \item L0, L1 -> L2: Here, we test the models' ability to compose skills from L0 and L1 tasks and perform higher level L2 tasks zero-shot and using fine-tuning.
\end{enumerate}

\section{Experiments}

\subsection{Baselines}

\begin{figure}[t!]
  \centering
  \includegraphics[width=1.0\linewidth]{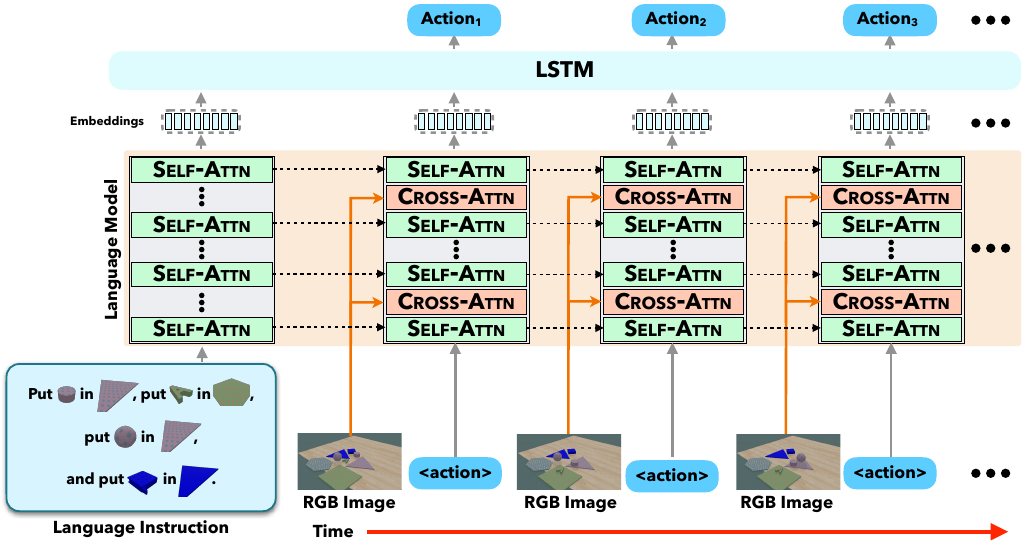}
  \vspace{-0.2cm}
  \caption{The StreamRoboLM model in contrast to state of the art models, \emph{e.g.}, RoboFlamingo (\emph{c.f.}, Fig.~1 in \cite{li2023vision}), can auto-regressively process videos as input, which helps for success in long-horizon tasks of ClevrSkills.}
  \label{fig:arch}
\end{figure}

For baselines, we evaluate open-source vision language policies that can take multi-modal prompts as inputs. We experiment with three different architectures: JAT~\cite{gallouedec2024jack} \rebuttal{and Octo~\cite{team2024octo}}, which accept image tokens in the context window of a transformer model, RoboFlamingo~\cite{li2023vision}, which uses cross attention to condition on the image embeddings generated from a vision encoder, and our own StreamRoboLM, which is based on the LRR model~\cite{bhattacharyya2023look} that continuously ingests video input during auto-regressive token generation. 

\textbf{JAT.} The Jack of All Trades (JAT)~\cite{gallouedec2024jack} model is an open-source generalist agent trained on a range of reinforcement learning and language and vision tasks. While JAT is trained on a large number of language only, vision-language and RL tasks, it can only perform one task at a time i.e., it can either model language or take image inputs to produce actions for RL tasks which means that none of the RL tasks can be specified using language. We modify JAT by simultaneously feeding text and image tokens so that the RL tasks can be conditioned on multi-modal prompts. We initialize from the pre-trained JAT model and fine-tune it on ClevrSkills tasks. 

\rebuttal{\textbf{Octo.} Octo is another open-source generalist policy trained on Open X-embodiment dataset~\cite{padalkar2023open}. The architecture is very similar to JAT with a transformer backbone and readout heads. The model is trained using a diffusion objective. The Octo architecture is geared towards enabling finetuning on tasks with different observation and action spaces. We leverage this to add additional observations for the ``first-person'' camera and prompt images used in the multimodal prompts. We refer the reader to the Octo~\cite{team2024octo} paper for further details on the model. We initialize from the pre-trained Octo model and fine-tune it on ClevrSkills tasks.}

\textbf{RoboFlamingo.} RoboFlamingo~\cite{li2023vision} takes open-source VLMs and augments them with an additional LSTM~\cite{hochreiter1997long} based policy head. 
The base VLM takes language and image inputs and produces an embedding that is then passed to the policy head, which in turn produces the next action. Since the base VLM is frozen, it basically acts as a ``prompt-processor'' which specifies the task to be performed by the LSTM policy. The model is trained on CALVIN dataset achieving good performance on the long-horizon tasks benchmark. We take the pre-trained RoboFlamingo model and further fine-tune it on ClevrSkills tasks.

\textbf{StreamRoboLM.} Different from RoboFlamingo, where the VLM can only reason over a single image at a time, we adapt an LRR~\cite{bhattacharyya2023look} based model that can auto-regressively take videos as input. Inspired by RoboFlamingo, we also attach an LSTM based policy head that takes token 
embeddings from the language model as input and produces the next action. 
Concretely, we use OPT1B~\cite{zhang2022opt} or Llama3.2 3B~\cite{dubey2024llama} as the base language model and a ViT~\cite{dosovitskiy2020image} as the vision encoder for the input images. The cross attention layers to condition on images and the LSTM based policy head are randomly initialized. To retain the language capabilities of the model, we use LoRA~\cite{hu2022lora} to fine-tune the language model while training all the parameters of the vision encoder and the policy head. \rebuttal{The architecture diagram can be seen in Figure~\ref{fig:arch}.}
Further details of the architectures are described in Appendix~\ref{app:baselines}. 

\begin{wrapfigure}{r}{0.65\textwidth}
    \begin{center}
      \centering
      \includegraphics[width=1.0\linewidth]{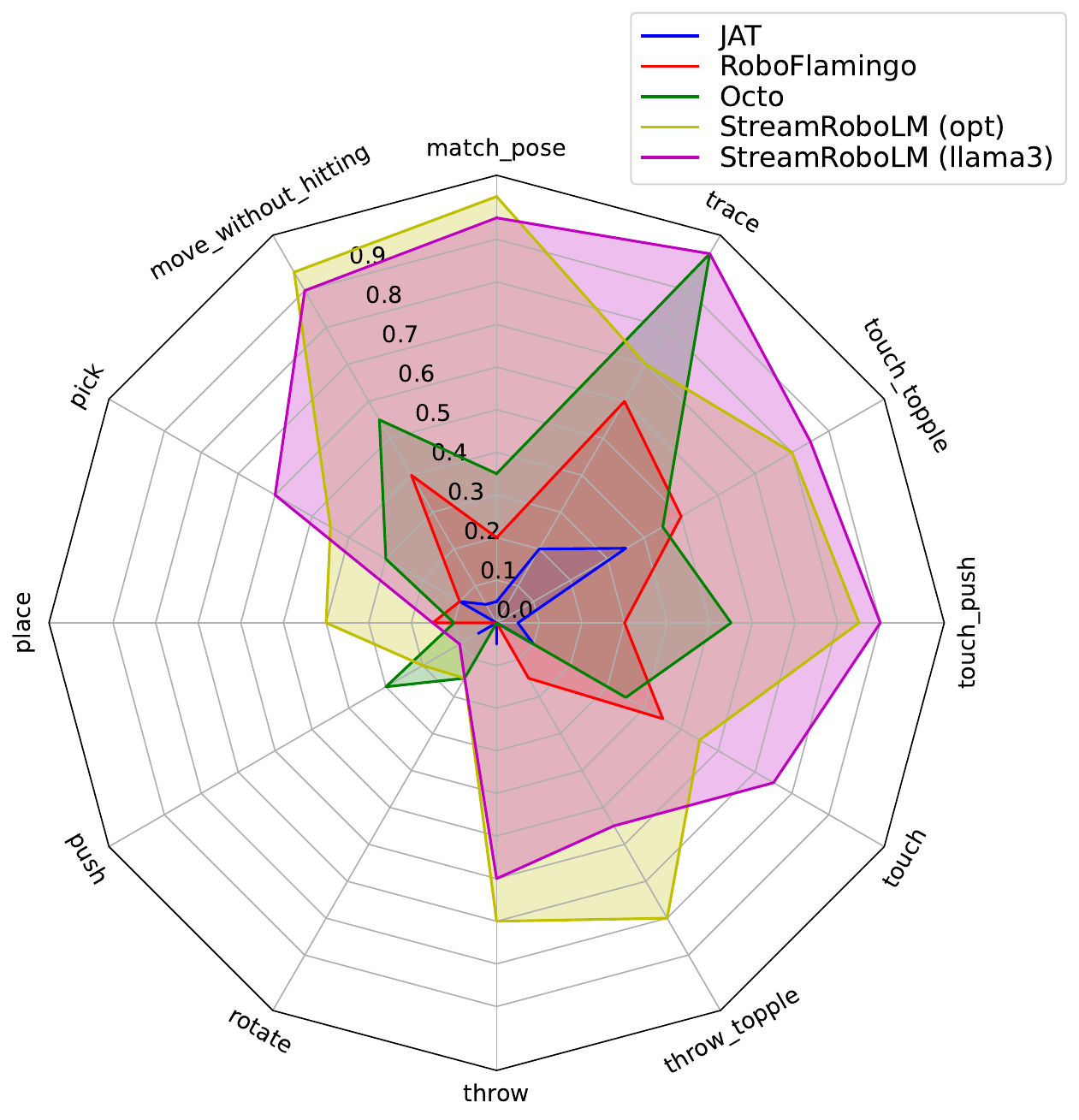}
    \end{center}
    \caption{Per task success rate on L0 tasks.}
    \label{fig:per_task_success}
\end{wrapfigure}

\subsection{Training and evaluation details} We use the open-source implementations for JAT, Octo and RoboFlamingo to evaluate the models on ClevrSkills. We initialize both the models from released checkpoints and fine-tune them on the ClevrSkills data on each task level separately. For StreamRoboLM, we start with OPT1B/Llama3.2 3B weights for the base LLM and ViT trained on ImageNet for the vision encoder. We initialize the cross-attention layers and the LSTM based policy head from scratch. We use LoRA~\cite{hu2022lora} while fine-tuning the LLM to retain the learned language capabilities while training all the parameters of the other modules. All the models were evaluated on 20 different seeds not seen at training time for each task in all the task levels. All the experiments were carried out on 4 NVIDIA A100 GPUs. Please refer to the Appendix~\ref{sec:training} for further details on training/evaluation hyperparameters.

\subsection{Results}

\begin{table*}[t]
  \centering
  \scriptsize
  \begin{tabularx}{1.0\textwidth}{Xc@{\hspace{0.3cm}}c@{\hspace{0.3cm}}c@{\hspace{0.3cm}}c@{\hspace{0.3cm}}c@{\hspace{0.3cm}}c@{\hspace{0.3cm}}c@{\hspace{0.3cm}}c@{\hspace{0.3cm}}}
    \toprule
    & \multicolumn{3}{c}{\textbf{Unseen seeds}} & \multicolumn{3}{c}{\textbf{Unseen objects/textures}} \\
    \cmidrule(lr){2-4}
    \cmidrule(lr){5-7}
    \textbf{Model} & \textbf{Success} & \textbf{Avg. Reward} & \textbf{Reward/Step} & \textbf{Success} & \textbf{Avg. Reward} & \textbf{Reward/Step}\\
    \midrule
    Oracle & 100.0 & 320.00 & 3.06 & 100.0 & 175.83 & 3.06 \\
    \rebuttal{JAT~\cite{gallouedec2024jack}} & 23.75 & 262.92 & 2.29 & 24.16 & 276.86 & 2.42\\
    \rebuttal{RoboFlamingo~\cite{li2023vision}} & 35.41 & 341.61 & 2.53 & 27.91 & 175.07 & 1.78 \\
    \rebuttal{Octo~\cite{team2024octo}} & 34.16 & 207.90 & 1.89 & 26.25 & 123.85 & 1.77 \\
    StreamRoboLM (Opt) & 62.91 & 223.84 & 2.74 & 41.66 & 329.69 & 2.87 \\
    \rebuttal{StreamRoboLM (Llama3)} & 62.50 & 198.94 & 2.65 & 55.41 & 215.52 & 2.65 \\
    \bottomrule
  \end{tabularx}
  \caption{Evaluation on L0 tasks.}
  \label{tab:l0_results}
\end{table*}

\textbf{Results on L0 tasks.} Results on the L0 tasks can be seen in Table~\ref{tab:l0_results}. 
It shows that all the models including JAT, RoboFlamingo, Octo and StreamRoboLM are able to perform some but not all L0 tasks. 
While the first three models only achieve success rates of $23.75\%$, $35.41\%$ and $34.16\%$, the StreamRoboLM (both, the OPT and Llama version) shows a decent performance overall of $62.91\%$ and $62.5\%$ success rate. 
We also show the task-wise success rate on L0 tasks in Figure~\ref{fig:per_task_success}. As we can see, some of the tasks such as rotate, push, place and pick are particularly difficult for all the models. Note that these tasks still require visual understanding of the scene and objects as the model needs to select the correct object to manipulate in the correct manner. 
The poor results can also be partially attributed to the strictness of the success criteria, 
since the models tend to obtain a decent average reward. 
We also show results of these models on unseen objects and textures to test their generalization capabilities on the same task with different objects and textures. As we can see, all models struggle to achieve similar performance as on seen object and textures, which shows that there is room for improvement. \rebuttal{Interestingly, JAT achieves similar performance on both seen and unseen objects and textures. 
This may be attributed to the smaller vision backbone which avoids overfitting.}

\begin{table*}[t]
  \centering
  \scriptsize
  \begin{tabularx}{\linewidth}{lcccccccccccc}
    \toprule
    & \multicolumn{6}{c}{\textbf{L1 Tasks}} & \multicolumn{6}{c}{\textbf{L2 Tasks}} \\
    \cmidrule(lr){2-7}
    \cmidrule(lr){8-13}
     & \multicolumn{3}{c}{\textbf{Zero-shot}} &  \multicolumn{3}{c}{\textbf{Fine-Tuning}} & \multicolumn{3}{c}{\textbf{Zero-shot}} & \multicolumn{3}{c}{\textbf{Fine-Tuning}} \\
    \cmidrule(lr){2-4}
    \cmidrule(lr){5-7}
    \cmidrule(lr){8-10}
    \cmidrule(lr){11-13}
    \textbf{Model} & \textbf{Suc.} & \textbf{AR} & \textbf{R/S} & \textbf{Suc.} & \textbf{AR} & \textbf{R/S} & \textbf{Suc.} & \textbf{AR} & \textbf{R/S} & \textbf{Suc.} & \textbf{AR} & \textbf{R/S} \\
    \midrule
    Oracle & 100.0 & 1027 & 5.59 & 100.0 & 1027 & 5.59 & 100.0 & 2583 & 9.12 & 100.0 & 2583 & 9.12 \\
    \rebuttal{JAT~\cite{gallouedec2024jack}} & 0.0 & 199 & 0.87 & 0.60 & 375 & 1.08 & 0.83 & 1344 & 2.15 & 0.0 & 1461 & 2.33 \\
    \rebuttal{RoboFlamingo~\cite{li2023vision}} & 0.0 & 268 & 0.79 & 1.00 & 452 & 1.29 & 0.0 & 1420 & 2.28 & 1.66 & 1493 & 2.39 \\
    \rebuttal{Octo~\cite{team2024octo}} & 0.33 & 326 & 0.94 & 5.00 & 469 & 1.45 & 0.83 & 1040 & 1.80 & 5.83 & 2106 & 3.53 \\
    StreamRoboLM (Opt) & 0.0 & 248 & 1.06 & 3.33 & 449 & 1.36 & 0.0 & 757 & 1.19 & 4.16 & 1047 & 1.71 \\
    \rebuttal{StreamRoboLM (Llama3)} & 0.0 & 352 & 1.01 & 3.33 & 497 & 1.44 & 0.0 & 1163 & 1.81 & 3.33 & 1566 & 2.57 \\
    \bottomrule
  \end{tabularx}
  \caption{Zero-shot generalization and fine-tuning results on L1 and L2 tasks. Here, \textit{Suc.} denotes Success rate, \textit{AR} denotes Avg. reward and \textit{R/S} denotes Reward per Step.}
  \label{tab:main_results}
\end{table*}

\textbf{Zero-shot generalization to L1 and L2 tasks.}
Zero-shot generalization results on L1 and L2 tasks are shown in Table~\ref{tab:main_results}. We train the models on L0 tasks and evaluate zero-shot on L1 and then fine-tune the models on L1 tasks and zero-shot evaluate resulting models on L2 tasks. This is the hardest instantiation of our benchmark as the tasks are not seen at training time and the model needs to compose the skills learned from training on L0 and L1 tasks to solve L1 and L2 tasks respectively. Unsurprisingly, we see that all the models struggle to generalize to these tasks in the zero-shot setting and only Octo~\cite{team2024octo} achieves non-zero success rate on any of the tasks. This can also be attributed to the significant complexity both in terms of length of episodes and compositional complexity of the tasks (see Fig.~\ref{fig:task_complexity}) in comparison to L0 tasks.

\textbf{Fine-tuning on L1 and L2 tasks.}
We also test these models by fine-tuning on L1 and L2 tasks. We found that despite training L1 and L2 tasks, these models struggle on both levels. As we can see in Figure~\ref{fig:task_complexity}, all these tasks are significantly longer than L0 tasks and require multiple successful executions of L0 skills ($\approx$ 9 on average for L1 and $\approx$ 11 on average for L2 tasks) for successful completion. As we also show in Figure~\ref{fig:task_comp}, the L1 and L2 tasks require more than simple stitching of L0 skills. %

Overall, the results show that even with reasonable performance on the L0 base skills, it is hard for current models to generalize to more complex tasks composed of these skills.

\rebuttal{Since ClevrSkills supports both multi-modal and language-only prompts, we also include experiments on language-only task specifications, which can be seen in Appendix~\ref{app:lang_only}.}

\section{Limitations and Future Work}
The main limitation is that our benchmark is fully simulated and building a real-world counterpart is the obvious future work.
Another direction for future work is the inclusion of more abstract and free-form tasks, such as playing tic-tac-toe, building structures like pyramids, houses, etc., aimed at evaluating long-range reasoning in robotics models, as well as the addition of multiple different embodiments (e.g. two-fingered grippers, dexterous grippers, or bi-manual robots). %

\section{Conclusion}
We present a benchmark for evaluating compositional understanding in robotics. To this end, we develop 33 tasks spread across 3 levels of compositional understanding. We benchmark state-of-art robotics models based on large vision language models (VLMs) and show that even after being trained on large amounts of internet and robotics data, these VLMs are unable to show good compositional generalization to new tasks. Overall, these results show that despite recent progress in both, VLMs and robotics, further research will be required for models to show compositional generalization capabilities in robotics.

\bibliographystyle{abbrvnat}
\bibliography{neurips_2024}

\begin{thebibliography}{50}
\providecommand{\natexlab}[1]{#1}
\providecommand{\url}[1]{\texttt{#1}}
\expandafter\ifx\csname urlstyle\endcsname\relax
  \providecommand{\doi}[1]{doi: #1}\else
  \providecommand{\doi}{doi: \begingroup \urlstyle{rm}\Url}\fi

\bibitem[Alayrac et~al.(2022)Alayrac, Donahue, Luc, Miech, Barr, Hasson, Lenc, Mensch, Millican, Reynolds, et~al.]{alayrac2022flamingo}
J.-B. Alayrac, J.~Donahue, P.~Luc, A.~Miech, I.~Barr, Y.~Hasson, K.~Lenc, A.~Mensch, K.~Millican, M.~Reynolds, et~al.
\newblock Flamingo: a visual language model for few-shot learning.
\newblock \emph{Advances in neural information processing systems}, 35:\penalty0 23716--23736, 2022.

\bibitem[Bahdanau et~al.(2020)Bahdanau, de~Vries, O'Donnell, Murty, Beaudoin, Bengio, and Courville]{bahdanau2020closure}
D.~Bahdanau, H.~de~Vries, T.~J. O'Donnell, S.~Murty, P.~Beaudoin, Y.~Bengio, and A.~Courville.
\newblock Closure: Assessing systematic generalization of clevr models, 2020.

\bibitem[Bao et~al.(2023)Bao, Xu, Qin, and Wang]{bao2023dexart}
C.~Bao, H.~Xu, Y.~Qin, and X.~Wang.
\newblock Dexart: Benchmarking generalizable dexterous manipulation with articulated objects.
\newblock In \emph{Proceedings of the IEEE/CVF Conference on Computer Vision and Pattern Recognition}, pages 21190--21200, 2023.

\bibitem[Bhattacharyya et~al.(2023)Bhattacharyya, Panchal, Pourreza, Lee, Madan, and Memisevic]{bhattacharyya2023look}
A.~Bhattacharyya, S.~Panchal, R.~Pourreza, M.~Lee, P.~Madan, and R.~Memisevic.
\newblock Look, remember and reason: Grounded reasoning in videos with language models.
\newblock In \emph{The Twelfth International Conference on Learning Representations}, 2023.

\bibitem[Brohan et~al.(2022)Brohan, Brown, Carbajal, Chebotar, Dabis, Finn, Gopalakrishnan, Hausman, Herzog, Hsu, et~al.]{brohan2022rt}
A.~Brohan, N.~Brown, J.~Carbajal, Y.~Chebotar, J.~Dabis, C.~Finn, K.~Gopalakrishnan, K.~Hausman, A.~Herzog, J.~Hsu, et~al.
\newblock Rt-1: Robotics transformer for real-world control at scale.
\newblock \emph{arXiv preprint arXiv:2212.06817}, 2022.

\bibitem[Brohan et~al.(2023)Brohan, Brown, Carbajal, Chebotar, Chen, Choromanski, Ding, Driess, Dubey, Finn, et~al.]{brohan2023rt}
A.~Brohan, N.~Brown, J.~Carbajal, Y.~Chebotar, X.~Chen, K.~Choromanski, T.~Ding, D.~Driess, A.~Dubey, C.~Finn, et~al.
\newblock Rt-2: Vision-language-action models transfer web knowledge to robotic control.
\newblock \emph{arXiv preprint arXiv:2307.15818}, 2023.

\bibitem[Cobbe et~al.(2021)Cobbe, Kosaraju, Bavarian, Chen, Jun, Kaiser, Plappert, Tworek, Hilton, Nakano, Hesse, and Schulman]{cobbe2021gsm8k}
K.~Cobbe, V.~Kosaraju, M.~Bavarian, M.~Chen, H.~Jun, L.~Kaiser, M.~Plappert, J.~Tworek, J.~Hilton, R.~Nakano, C.~Hesse, and J.~Schulman.
\newblock Training verifiers to solve math word problems.
\newblock \emph{arXiv preprint arXiv:2110.14168}, 2021.

\bibitem[Dosovitskiy et~al.(2020)Dosovitskiy, Beyer, Kolesnikov, Weissenborn, Zhai, Unterthiner, Dehghani, Minderer, Heigold, Gelly, et~al.]{dosovitskiy2020image}
A.~Dosovitskiy, L.~Beyer, A.~Kolesnikov, D.~Weissenborn, X.~Zhai, T.~Unterthiner, M.~Dehghani, M.~Minderer, G.~Heigold, S.~Gelly, et~al.
\newblock An image is worth 16x16 words: Transformers for image recognition at scale.
\newblock \emph{arXiv preprint arXiv:2010.11929}, 2020.

\bibitem[Dubey et~al.(2024)Dubey, Jauhri, Pandey, Kadian, Al-Dahle, Letman, Mathur, Schelten, Yang, Fan, et~al.]{dubey2024llama}
A.~Dubey, A.~Jauhri, A.~Pandey, A.~Kadian, A.~Al-Dahle, A.~Letman, A.~Mathur, A.~Schelten, A.~Yang, A.~Fan, et~al.
\newblock The llama 3 herd of models.
\newblock \emph{arXiv preprint arXiv:2407.21783}, 2024.

\bibitem[Ebert et~al.(2021)Ebert, Yang, Schmeckpeper, Bucher, Georgakis, Daniilidis, Finn, and Levine]{ebert2021bridge}
F.~Ebert, Y.~Yang, K.~Schmeckpeper, B.~Bucher, G.~Georgakis, K.~Daniilidis, C.~Finn, and S.~Levine.
\newblock Bridge data: Boosting generalization of robotic skills with cross-domain datasets.
\newblock \emph{arXiv preprint arXiv:2109.13396}, 2021.

\bibitem[Fang et~al.(2023)Fang, Fang, Tang, Liu, Wang, Zhu, and Lu]{fang2023rh20t}
H.-S. Fang, H.~Fang, Z.~Tang, J.~Liu, J.~Wang, H.~Zhu, and C.~Lu.
\newblock Rh20t: A robotic dataset for learning diverse skills in one-shot.
\newblock \emph{arXiv preprint arXiv:2307.00595}, 2023.

\bibitem[Gallou{\'e}dec et~al.(2024)Gallou{\'e}dec, Beeching, Romac, and Dellandr{\'e}a]{gallouedec2024jack}
Q.~Gallou{\'e}dec, E.~Beeching, C.~Romac, and E.~Dellandr{\'e}a.
\newblock Jack of all trades, master of some, a multi-purpose transformer agent.
\newblock \emph{arXiv preprint arXiv:2402.09844}, 2024.

\bibitem[Gebru et~al.(2021)Gebru, Morgenstern, Vecchione, Vaughan, Wallach, Iii, and Crawford]{gebru2021datasheets}
T.~Gebru, J.~Morgenstern, B.~Vecchione, J.~W. Vaughan, H.~Wallach, H.~D. Iii, and K.~Crawford.
\newblock Datasheets for datasets.
\newblock \emph{Communications of the ACM}, 64\penalty0 (12):\penalty0 86--92, 2021.

\bibitem[Gong et~al.(2023)Gong, Huang, Zhao, Geng, Gao, Wu, Ai, Zhou, Terzopoulos, Zhu, et~al.]{gong2023arnold}
R.~Gong, J.~Huang, Y.~Zhao, H.~Geng, X.~Gao, Q.~Wu, W.~Ai, Z.~Zhou, D.~Terzopoulos, S.-C. Zhu, et~al.
\newblock Arnold: A benchmark for language-grounded task learning with continuous states in realistic 3d scenes.
\newblock In \emph{Proceedings of the IEEE/CVF International Conference on Computer Vision}, pages 20483--20495, 2023.

\bibitem[Goyal et~al.(2017)Goyal, Ebrahimi~Kahou, Michalski, Materzynska, Westphal, Kim, Haenel, Fruend, Yianilos, Mueller-Freitag, et~al.]{goyal2017something}
R.~Goyal, S.~Ebrahimi~Kahou, V.~Michalski, J.~Materzynska, S.~Westphal, H.~Kim, V.~Haenel, I.~Fruend, P.~Yianilos, M.~Mueller-Freitag, et~al.
\newblock The" something something" video database for learning and evaluating visual common sense.
\newblock In \emph{Proceedings of the IEEE international conference on computer vision}, pages 5842--5850, 2017.

\bibitem[Gu et~al.(2023)Gu, Xiang, Li, Ling, Liu, Mu, Tang, Tao, Wei, Yao, et~al.]{gu2023maniskill2}
J.~Gu, F.~Xiang, X.~Li, Z.~Ling, X.~Liu, T.~Mu, Y.~Tang, S.~Tao, X.~Wei, Y.~Yao, et~al.
\newblock Maniskill2: A unified benchmark for generalizable manipulation skills.
\newblock In \emph{The Eleventh International Conference on Learning Representations}, 2023.

\bibitem[Hao Su's~Lab(2024)]{mplib}
U.~Hao Su's~Lab.
\newblock {MPlib}: a lightweight motion planning library.
\newblock \url{https://github.com/haosulab/MPlib}, 2024.

\bibitem[Hochreiter and Schmidhuber(1997)]{hochreiter1997long}
S.~Hochreiter and J.~Schmidhuber.
\newblock Long short-term memory.
\newblock \emph{Neural computation}, 9\penalty0 (8):\penalty0 1735--1780, 1997.

\bibitem[Hu et~al.(2022)Hu, Shen, Wallis, Allen-Zhu, Li, Wang, Wang, and Chen]{hu2022lora}
E.~J. Hu, Y.~Shen, P.~Wallis, Z.~Allen-Zhu, Y.~Li, S.~Wang, L.~Wang, and W.~Chen.
\newblock Lo{RA}: Low-rank adaptation of large language models.
\newblock In \emph{International Conference on Learning Representations}, 2022.

\bibitem[Jiang et~al.(2023)Jiang, Gupta, Zhang, Wang, Dou, Chen, Fei-Fei, Anandkumar, Zhu, and Fan]{jiang2023vima}
Y.~Jiang, A.~Gupta, Z.~Zhang, G.~Wang, Y.~Dou, Y.~Chen, L.~Fei-Fei, A.~Anandkumar, Y.~Zhu, and L.~Fan.
\newblock Vima: General robot manipulation with multimodal prompts.
\newblock In \emph{Fortieth International Conference on Machine Learning}, 2023.

\bibitem[Johnson et~al.(2017)Johnson, Hariharan, van~der Maaten, Fei-Fei, Zitnick, and Girshick]{johnson2017clevr}
J.~Johnson, B.~Hariharan, L.~van~der Maaten, L.~Fei-Fei, C.~L. Zitnick, and R.~Girshick.
\newblock Clevr: A diagnostic dataset for compositional language and elementary visual reasoning.
\newblock In \emph{CVPR}, 2017.

\bibitem[Kahneman(2011)]{kahneman2011thinking}
D.~Kahneman.
\newblock \emph{Thinking, fast and slow}.
\newblock Farrar, Straus and Giroux, New York, 2011.
\newblock ISBN 9780374275631 0374275637.
\newblock URL \url{https://www.amazon.de/Thinking-Fast-Slow-Daniel-Kahneman/dp/0374275637/ref=wl_it_dp_o_pdT1_nS_nC?ie=UTF8&colid=151193SNGKJT9&coliid=I3OCESLZCVDFL7}.

\bibitem[Kolve et~al.(2017)Kolve, Mottaghi, Han, VanderBilt, Weihs, Herrasti, Deitke, Ehsani, Gordon, Zhu, et~al.]{kolve2017ai2}
E.~Kolve, R.~Mottaghi, W.~Han, E.~VanderBilt, L.~Weihs, A.~Herrasti, M.~Deitke, K.~Ehsani, D.~Gordon, Y.~Zhu, et~al.
\newblock Ai2-thor: An interactive 3d environment for visual ai.
\newblock \emph{arXiv preprint arXiv:1712.05474}, 2017.

\bibitem[LaValle.(1998)]{rrt}
S.~M. LaValle.
\newblock Rapidly-exploring random trees: A new tool for path planning.
\newblock \emph{TR 98-11, Computer Science Dept., Iowa State University, October 1998}, 1998.

\bibitem[Lerer et~al.(2016)Lerer, Gross, and Fergus]{pmlr-v48-lerer16}
A.~Lerer, S.~Gross, and R.~Fergus.
\newblock Learning physical intuition of block towers by example.
\newblock In M.~F. Balcan and K.~Q. Weinberger, editors, \emph{Proceedings of The 33rd International Conference on Machine Learning}, volume~48 of \emph{Proceedings of Machine Learning Research}, pages 430--438, New York, New York, USA, 20--22 Jun 2016. PMLR.
\newblock URL \url{https://proceedings.mlr.press/v48/lerer16.html}.

\bibitem[Li et~al.(2021)Li, Xia, Mart{\'\i}n-Mart{\'\i}n, Lingelbach, Srivastava, Shen, Vainio, Gokmen, Dharan, Jain, et~al.]{li2021igibson}
C.~Li, F.~Xia, R.~Mart{\'\i}n-Mart{\'\i}n, M.~Lingelbach, S.~Srivastava, B.~Shen, K.~Vainio, C.~Gokmen, G.~Dharan, T.~Jain, et~al.
\newblock igibson 2.0: Object-centric simulation for robot learning of everyday household tasks.
\newblock \emph{arXiv preprint arXiv:2108.03272}, 2021.

\bibitem[Li et~al.(2023)Li, Zhang, Wong, Gokmen, Srivastava, Mart{\'\i}n-Mart{\'\i}n, Wang, Levine, Lingelbach, Sun, et~al.]{li2023behavior}
C.~Li, R.~Zhang, J.~Wong, C.~Gokmen, S.~Srivastava, R.~Mart{\'\i}n-Mart{\'\i}n, C.~Wang, G.~Levine, M.~Lingelbach, J.~Sun, et~al.
\newblock Behavior-1k: A benchmark for embodied ai with 1,000 everyday activities and realistic simulation.
\newblock In \emph{Conference on Robot Learning}, pages 80--93. PMLR, 2023.

\bibitem[Li et~al.(2024)Li, Liu, Zhang, Yu, Xu, Wu, Cheang, Jing, Zhang, Liu, et~al.]{li2023vision}
X.~Li, M.~Liu, H.~Zhang, C.~Yu, J.~Xu, H.~Wu, C.~Cheang, Y.~Jing, W.~Zhang, H.~Liu, et~al.
\newblock Vision-language foundation models as effective robot imitators.
\newblock \emph{The Twelfth International Conference on Learning Representations}, 2024.

\bibitem[Liu et~al.(2023)Liu, Li, Wu, and Lee]{liu2023llava}
H.~Liu, C.~Li, Q.~Wu, and Y.~J. Lee.
\newblock Visual instruction tuning.
\newblock In \emph{NeurIPS}, 2023.

\bibitem[Luo et~al.(2024)Luo, Xu, Liu, Tan, Lin, Wu, Abbeel, and Levine]{luo2024fmb}
J.~Luo, C.~Xu, F.~Liu, L.~Tan, Z.~Lin, J.~Wu, P.~Abbeel, and S.~Levine.
\newblock Fmb: a functional manipulation benchmark for generalizable robotic learning.
\newblock \emph{arXiv preprint arXiv:2401.08553}, 2024.

\bibitem[Mandlekar et~al.(2018)Mandlekar, Zhu, Garg, Booher, Spero, Tung, Gao, Emmons, Gupta, Orbay, et~al.]{mandlekar2018roboturk}
A.~Mandlekar, Y.~Zhu, A.~Garg, J.~Booher, M.~Spero, A.~Tung, J.~Gao, J.~Emmons, A.~Gupta, E.~Orbay, et~al.
\newblock Roboturk: A crowdsourcing platform for robotic skill learning through imitation.
\newblock In \emph{Conference on Robot Learning}, pages 879--893. PMLR, 2018.

\bibitem[Mees et~al.(2022)Mees, Hermann, Rosete-Beas, and Burgard]{mees2022calvin}
O.~Mees, L.~Hermann, E.~Rosete-Beas, and W.~Burgard.
\newblock Calvin: A benchmark for language-conditioned policy learning for long-horizon robot manipulation tasks.
\newblock \emph{IEEE Robotics and Automation Letters}, 7\penalty0 (3):\penalty0 7327--7334, 2022.

\bibitem[Padalkar et~al.(2023)Padalkar, Pooley, Jain, Bewley, Herzog, Irpan, Khazatsky, Rai, Singh, Brohan, et~al.]{padalkar2023open}
A.~Padalkar, A.~Pooley, A.~Jain, A.~Bewley, A.~Herzog, A.~Irpan, A.~Khazatsky, A.~Rai, A.~Singh, A.~Brohan, et~al.
\newblock Open x-embodiment: Robotic learning datasets and rt-x models.
\newblock \emph{arXiv preprint arXiv:2310.08864}, 2023.

\bibitem[Paszke et~al.(2019)Paszke, Gross, Massa, Lerer, Bradbury, Chanan, Killeen, Lin, Gimelshein, Antiga, et~al.]{paszke2019pytorch}
A.~Paszke, S.~Gross, F.~Massa, A.~Lerer, J.~Bradbury, G.~Chanan, T.~Killeen, Z.~Lin, N.~Gimelshein, L.~Antiga, et~al.
\newblock Pytorch: An imperative style, high-performance deep learning library.
\newblock \emph{Advances in neural information processing systems}, 32, 2019.

\bibitem[Radford et~al.(2019)Radford, Wu, Child, Luan, Amodei, Sutskever, et~al.]{radford2019language}
A.~Radford, J.~Wu, R.~Child, D.~Luan, D.~Amodei, I.~Sutskever, et~al.
\newblock Language models are unsupervised multitask learners.
\newblock \emph{OpenAI blog}, 1\penalty0 (8):\penalty0 9, 2019.

\bibitem[Reed et~al.(2022)Reed, Zolna, Parisotto, Colmenarejo, Novikov, Barth-Maron, Gimenez, Sulsky, Kay, Springenberg, et~al.]{reed2022generalist}
S.~Reed, K.~Zolna, E.~Parisotto, S.~G. Colmenarejo, A.~Novikov, G.~Barth-Maron, M.~Gimenez, Y.~Sulsky, J.~Kay, J.~T. Springenberg, et~al.
\newblock A generalist agent.
\newblock \emph{arXiv preprint arXiv:2205.06175}, 2022.

\bibitem[Riveland and Pouget(2024)]{riveland2024natural}
R.~Riveland and A.~Pouget.
\newblock Natural language instructions induce compositional generalization in networks of neurons.
\newblock \emph{Nature Neuroscience}, pages 1--12, 2024.

\bibitem[Ross et~al.(2011)Ross, Gordon, and Bagnell]{ross2011reduction}
S.~Ross, G.~Gordon, and D.~Bagnell.
\newblock A reduction of imitation learning and structured prediction to no-regret online learning.
\newblock In \emph{Proceedings of the fourteenth international conference on artificial intelligence and statistics}, pages 627--635. JMLR Workshop and Conference Proceedings, 2011.

\bibitem[Shridhar et~al.(2022)Shridhar, Manuelli, and Fox]{shridhar2022cliport}
M.~Shridhar, L.~Manuelli, and D.~Fox.
\newblock Cliport: What and where pathways for robotic manipulation.
\newblock In \emph{Conference on robot learning}, pages 894--906. PMLR, 2022.

\bibitem[Shridhar et~al.(2023)Shridhar, Manuelli, and Fox]{shridhar2023perceiver}
M.~Shridhar, L.~Manuelli, and D.~Fox.
\newblock Perceiver-actor: A multi-task transformer for robotic manipulation.
\newblock In \emph{Conference on Robot Learning}, pages 785--799. PMLR, 2023.

\bibitem[{\c{S}}ucan et~al.(2012){\c{S}}ucan, Moll, and Kavraki]{ompl}
I.~A. {\c{S}}ucan, M.~Moll, and L.~E. Kavraki.
\newblock The {O}pen {M}otion {P}lanning {L}ibrary.
\newblock \emph{{IEEE} Robotics \& Automation Magazine}, 19\penalty0 (4):\penalty0 72--82, December 2012.
\newblock \doi{10.1109/MRA.2012.2205651}.
\newblock \url{https://ompl.kavrakilab.org}.

\bibitem[Szot et~al.(2021)Szot, Clegg, Undersander, Wijmans, Zhao, Turner, Maestre, Mukadam, Chaplot, Maksymets, et~al.]{szot2021habitat}
A.~Szot, A.~Clegg, E.~Undersander, E.~Wijmans, Y.~Zhao, J.~Turner, N.~Maestre, M.~Mukadam, D.~S. Chaplot, O.~Maksymets, et~al.
\newblock Habitat 2.0: Training home assistants to rearrange their habitat.
\newblock \emph{Advances in neural information processing systems}, 34:\penalty0 251--266, 2021.

\bibitem[Team et~al.(2024)Team, Ghosh, Walke, Pertsch, Black, Mees, Dasari, Hejna, Kreiman, Xu, et~al.]{team2024octo}
O.~M. Team, D.~Ghosh, H.~Walke, K.~Pertsch, K.~Black, O.~Mees, S.~Dasari, J.~Hejna, T.~Kreiman, C.~Xu, et~al.
\newblock Octo: An open-source generalist robot policy.
\newblock \emph{arXiv preprint arXiv:2405.12213}, 2024.

\bibitem[Weston et~al.(2015)Weston, Bordes, Chopra, Rush, van Merriënboer, Joulin, and Mikolov]{weston2015aicomplete}
J.~Weston, A.~Bordes, S.~Chopra, A.~M. Rush, B.~van Merriënboer, A.~Joulin, and T.~Mikolov.
\newblock Towards ai-complete question answering: A set of prerequisite toy tasks, 2015.

\bibitem[Winograd(1972)]{winograd1972understanding}
T.~Winograd.
\newblock \emph{Understanding Natural Language}.
\newblock Academic Press, 1972.
\newblock ISBN 9780127597508.
\newblock URL \url{https://books.google.ca/books?id=-FxQAAAAMAAJ}.

\bibitem[Wu et~al.(2016)Wu, Lim, Zhang, Tenenbaum, and Freeman]{phys101}
J.~Wu, J.~J. Lim, H.~Zhang, J.~B. Tenenbaum, and W.~T. Freeman.
\newblock Physics 101: Learning physical object properties from unlabeled videos.
\newblock In \emph{British Machine Vision Conference}, 2016.

\bibitem[Yi* et~al.(2020)Yi*, Gan*, Li, Kohli, Wu, Torralba, and Tenenbaum]{Yi*2020CLEVRER:}
K.~Yi*, C.~Gan*, Y.~Li, P.~Kohli, J.~Wu, A.~Torralba, and J.~B. Tenenbaum.
\newblock Clevrer: Collision events for video representation and reasoning.
\newblock In \emph{International Conference on Learning Representations}, 2020.
\newblock URL \url{https://openreview.net/forum?id=HkxYzANYDB}.

\bibitem[Yu et~al.(2024)Yu, Hao, Wang, Liu, Liu, Mu, You, Yan, and Lu]{yu2024manipose}
Q.~Yu, C.~Hao, J.~Wang, W.~Liu, L.~Liu, Y.~Mu, Y.~You, H.~Yan, and C.~Lu.
\newblock Manipose: A comprehensive benchmark for pose-aware object manipulation in robotics.
\newblock \emph{arXiv preprint arXiv:2403.13365}, 2024.

\bibitem[Yu et~al.(2020)Yu, Quillen, He, Julian, Hausman, Finn, and Levine]{yu2020meta}
T.~Yu, D.~Quillen, Z.~He, R.~Julian, K.~Hausman, C.~Finn, and S.~Levine.
\newblock Meta-world: A benchmark and evaluation for multi-task and meta reinforcement learning.
\newblock In \emph{Conference on robot learning}, pages 1094--1100. PMLR, 2020.

\bibitem[Zhang et~al.(2022)Zhang, Roller, Goyal, Artetxe, Chen, Chen, Dewan, Diab, Li, Lin, et~al.]{zhang2022opt}
S.~Zhang, S.~Roller, N.~Goyal, M.~Artetxe, M.~Chen, S.~Chen, C.~Dewan, M.~Diab, X.~Li, X.~V. Lin, et~al.
\newblock Opt: Open pre-trained transformer language models.
\newblock \emph{arXiv preprint arXiv:2205.01068}, 2022.

\end{thebibliography}
\clearpage

\appendix

\section{ClevrSkills Task Suite}
\label{app:tasks}

In this section, we present a detailed description of all the tasks included in the ClevrSkills task suite. The task suite comprises of 33 different tasks over 3 different levels of compositionality. 

\subsection{L0: Simple Tasks}
The tasks included in L0 are designed to teach the robot basic motor/manipulation skills. The skills can then be composed in various ways to achieve more interesting tasks in level L1 and level L2 of our benchmark. Below is the list of all the tasks in this level.

\begin{enumerate}

    \item \textbf{Match pose:} 
    \begin{center}
       \includegraphics[width=0.9\linewidth]{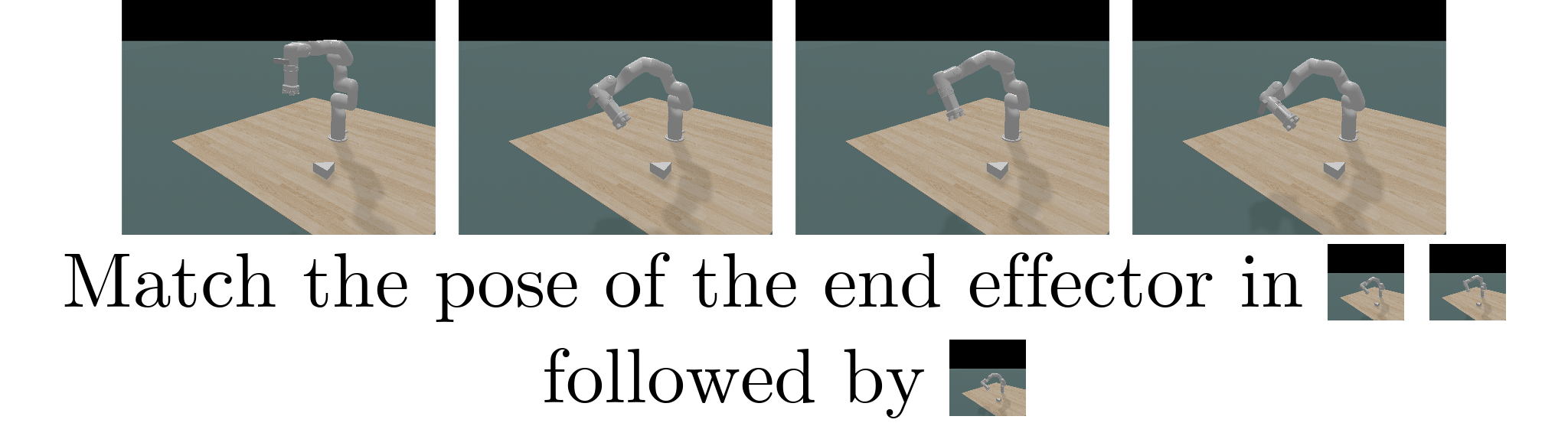}
    \end{center}
    \begin{itemize}
        \item \textbf{Prompts:} 
        \begin{enumerate}
            \item "Match the pose of the end effector in \{ks:keystep\_1\}, \{ks:keystep\_2\} followed by \{ks:keystep\_3\}".
        \end{enumerate}
        \item \textbf{Description:} The image placeholder \{ks:keystep\_1\} is the goal image showing the pose of the robot that it needs to achieve. The number of goals can vary in which case it needs to achieve the goal poses in the order in which the images are shown. The task allows the robot to learn to move the arm to achieve the required pose.
        \item \textbf{Success Criteria:} The combined error in both the position and the rotation of the pose of the end-effector should be less than $0.05$.
    \end{itemize}

    \item \textbf{Move without hitting:} 
    \begin{center}
        \includegraphics[width=0.9\linewidth]{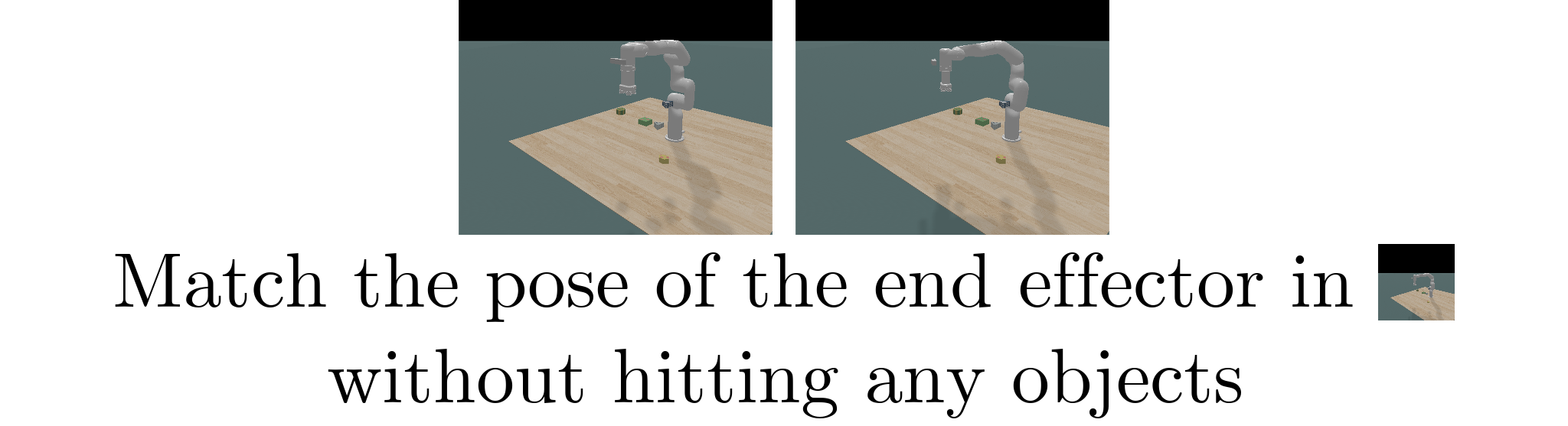}
    \end{center}
    \begin{itemize}
        \item \textbf{Prompts:} 
        \begin{enumerate}
            \item Match the pose of the end effector in \{ks:keystep\_1\} without hitting any objects.
        \end{enumerate}
        \item \textbf{Description:} Similar to 1, but with an additional constraint of avoiding objects suspended in air. The number of obstacles may vary from 1 to 5.
        \item \textbf{Success Criteria:} The combined error in both the position and the rotation of the pose of the end-effector should be less than $0.05$ and none of the obstacles are touched.
    \end{itemize}

    \item \textbf{Pick:} 
    \begin{center}
        \includegraphics[width=0.9\linewidth]{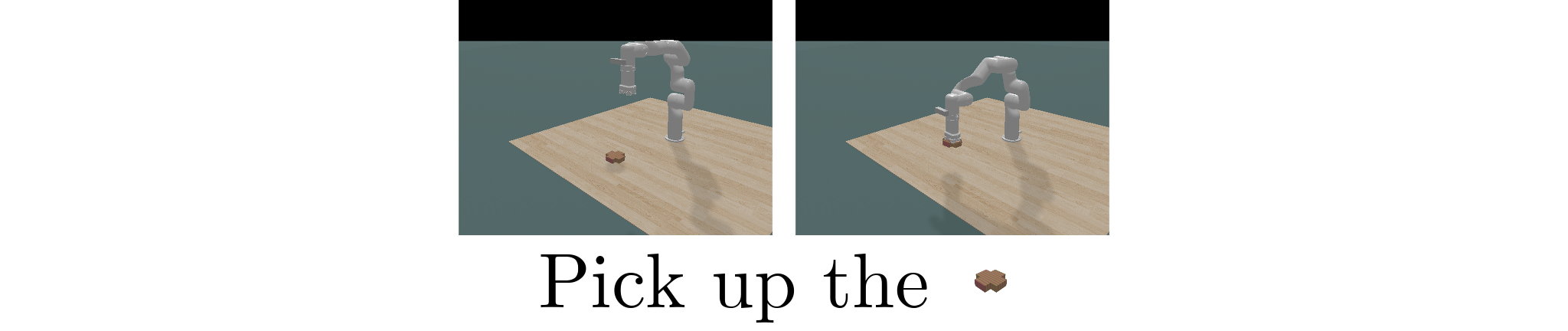}
    \end{center}
    \begin{itemize}
        \item \textbf{Prompts:} 
        \begin{enumerate}
            \item Pick up the \{obj:object\}.
            \item Grab the \{obj:object\}.
            \item Lift the \{obj:object\}.
            \item Pick up the object with \{tex:object\} texture.
            \item Grab the object with \{tex:object\} texture.
            \item Lift the object with \{tex:object\} texture.
        \end{enumerate}
        \item \textbf{Description:} The robot is required to pick the object shown in the image placeholder \{obj:object\} or the object having the texture shown in \{tex:object\}. To make it more difficult, we also introduce distractor objects which may range from 0 to 3 objects.
        \item \textbf{Success Criteria:} The specified object is picked i.e. attached to the end-effector and not touching the ground.
    \end{itemize}

    \item \textbf{Place:} 
    \begin{center}
        \includegraphics[width=0.9\linewidth]{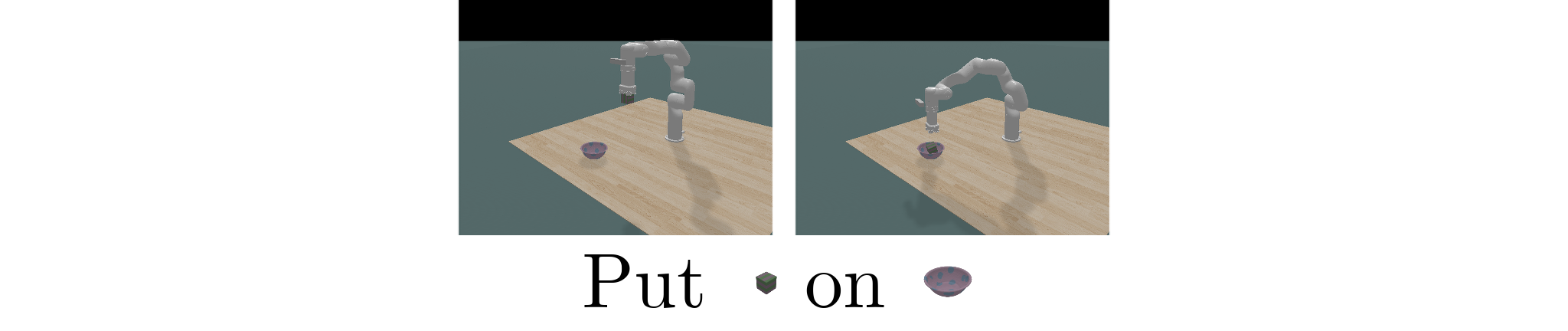}
    \end{center}
    \begin{itemize}
        \item \textbf{Prompts:} 
        \begin{enumerate}
            \item Put \{obj:object\}$_{1}$ on \{obj:object\}$_{2}$.
            \item Put object with {tex:object}$_{1}$ texture on object with {tex:object}$_{2}$ texture.
        \end{enumerate}
        \item \textbf{Description:} The agent starts off with an object (\{obj:object\}$_{1}$) attached to the end-effector and is tasked to place it on the specified object as shown in the placeholder \{obj:object\}$_{2}$. 
        \item \textbf{Success Criteria:} The object in end-effector is placed on the specified object and is not touching either the end-effector or the ground.
    \end{itemize}

    \item \textbf{Push:} 
    \begin{center}
        \includegraphics[width=0.9\linewidth]{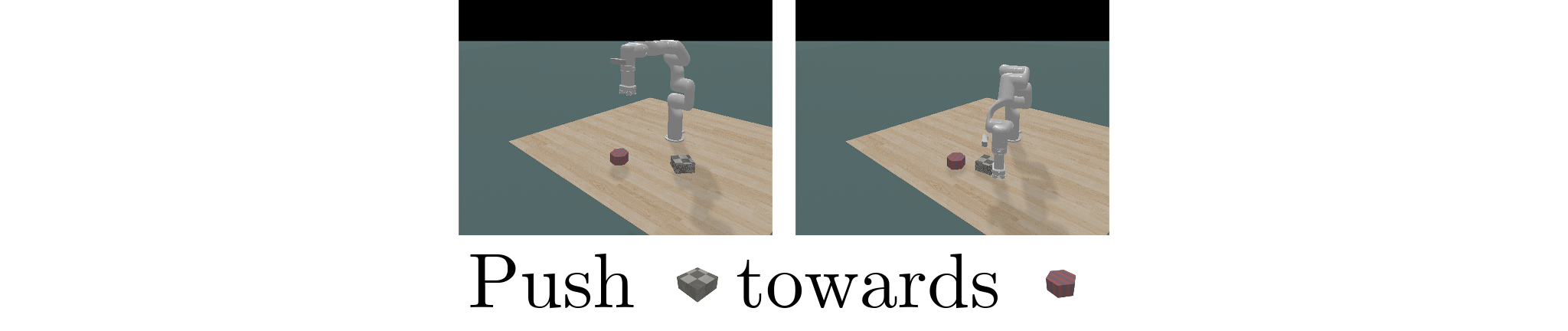}
    \end{center}
    \begin{itemize}
        \item \textbf{Prompts:} 
        \begin{enumerate}
            \item Push \{obj:object\}$_{1}$ towards \{obj:object\}$_{2}$.
            \item Push object with \{tex:object\}$_{1}$ texture towards object with \{tex:object\}$_{2}$ texture.
        \end{enumerate}
        \item \textbf{Description:} The agent is tasked to push the specified object \{obj:object\}$_{1}$ towards another object \{obj:object\}$_{2}$. 
        \item \textbf{Success Criteria:} The target object is pushed in the direction of goal object $\pm45$ degrees and the distance between target and goal object should reduce by $30\%$.
    \end{itemize}

    \item \textbf{Rotate:} 
    \begin{center}
        \includegraphics[width=0.9\linewidth]{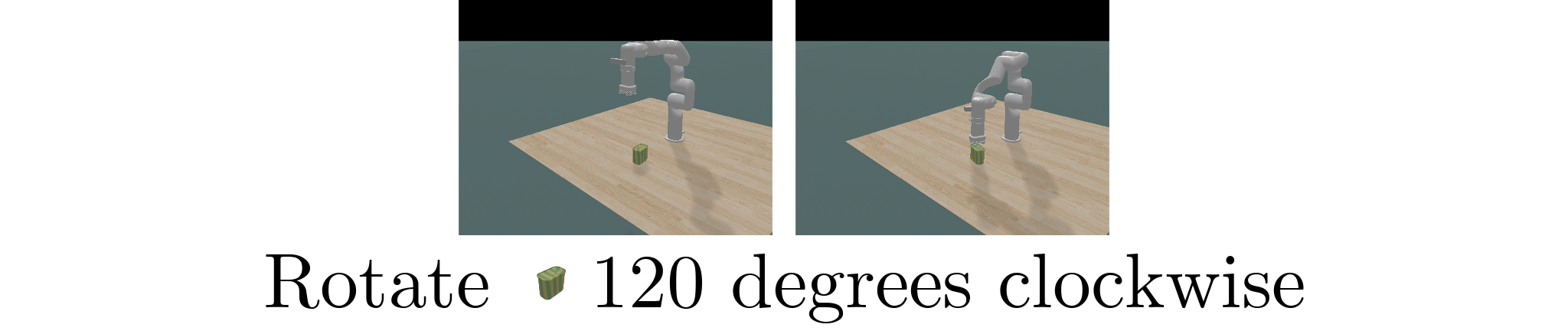}
    \end{center}
    \begin{itemize}
        \item \textbf{Prompts:} 
        \begin{enumerate}
            \item Rotate \{obj:object\}$_{1}$ \{angles\} degrees \{direction\}.
            \item Rotate object with \{tex:object\}$_{1}$ \{angles\} degrees \{direction\}.
        \end{enumerate}
        \item \textbf{Description:} The placeholder \{obj:object\}$_{1}$ specifies the object to be rotated by \{angles\} in \{direction\}. The angles can take values of $30$, $60$, $90$, $120$, and $150$ whereas the direction can take values of \textit{clockwise} or \textit{anti-clockwise}. There may be multiple objects that need to be rotated in order.
        \item \textbf{Success Criteria:} The specified object is rotated in the correct direction within $5$ degrees of the specified angle. The position of the object should not change more than $5$cm.
    \end{itemize}

    \item \textbf{Throw:} 
    \begin{center}
        \includegraphics[width=0.9\linewidth]{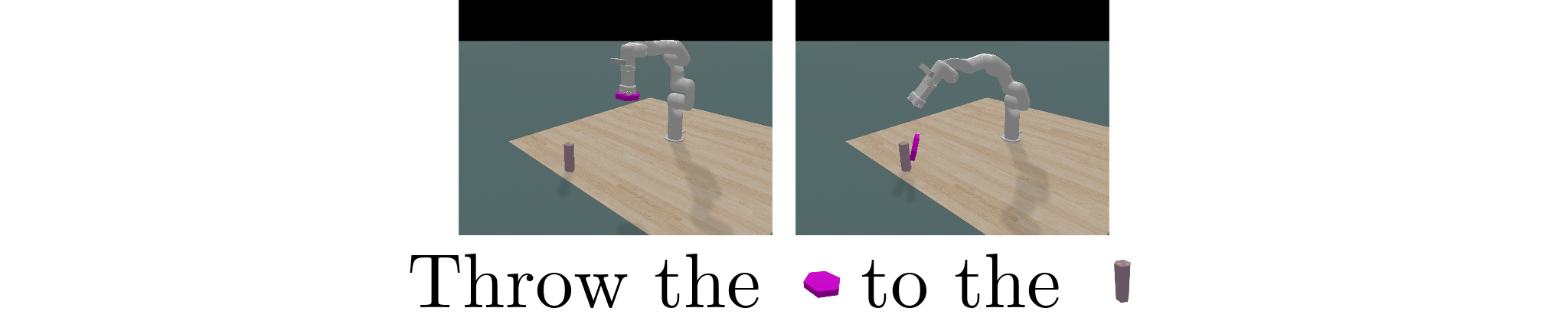}
    \end{center}
    \begin{itemize}
        \item \textbf{Prompts:} 
        \begin{enumerate}
            \item Throw \{obj:object\}$_{1}$ to \{obj:object\}$_{2}$.
            \item Hit \{obj:object\}$_{2}$ with \{obj:object\}$_{1}$.
        \end{enumerate}
        \item \textbf{Description:} The agent is tasked to throw \{obj:object\}$_{1}$ to \{obj:object\}$_{2}$. \{obj:object\}$_{1}$ is initialized at the end-effector so the robot does not need to first pick it up. The episode ends as soon as the target object is touched.
        \item \textbf{Success Criteria:} The thrown object must touch the specified goal object. 
    \end{itemize}

    \item \textbf{Throw topple:} 
    \begin{center}
        \includegraphics[width=0.9\linewidth]{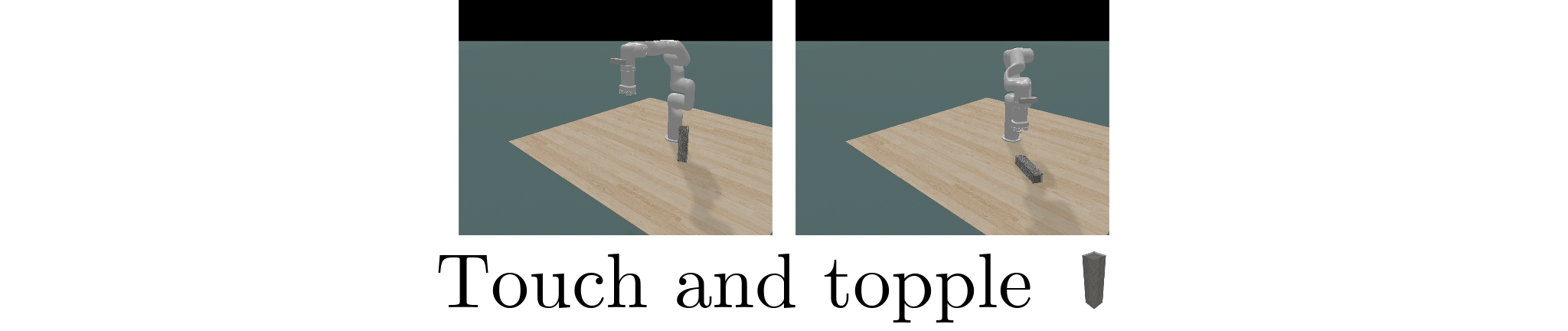}
    \end{center}
    \begin{itemize}
        \item \textbf{Prompts:} 
        \begin{enumerate}
            \item Throw \{obj:object\}$_{1}$ to \{obj:object\}$_{2}$ such that \{obj:object\}$_{2}$ falls over.
            \item Hit \{obj:object\}$_{2}$ with \{obj:object\}$_{1}$ such that \{obj:object\}$_{2}$ falls over.
        \end{enumerate}
        \item \textbf{Description:} Similar to 7, with the additional constraint that the target object must topple over. 
        \item \textbf{Success Criteria:} Same a 7 but the target object must topple over such that there is a more than $45$ degree change in the vertical axis of the object.
    \end{itemize}

    \item \textbf{Touch:} 
    \begin{center}
        \includegraphics[width=0.9\linewidth]{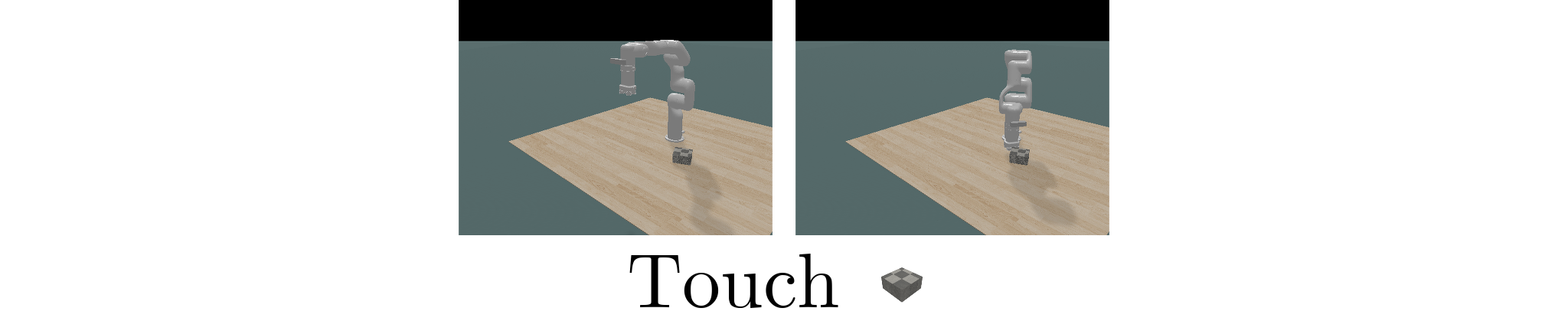}
    \end{center}
    \begin{itemize}
        \item \textbf{Prompts:} 
        \begin{enumerate}
            \item Touch \{obj:object\}$_{1}$.
        \end{enumerate}
        \item \textbf{Description:} The agent is tasked to gently touch a specified object without moving it. The task is supposed to teach the agent to control the force with which it carries out the task.
        \item \textbf{Success Criteria:} The specified object must be touched without moving it more than $3$cm. It must also not be grasped at any point.
    \end{itemize}

    \item \textbf{Touch push:} 
    \begin{center}
        \includegraphics[width=0.9\linewidth]{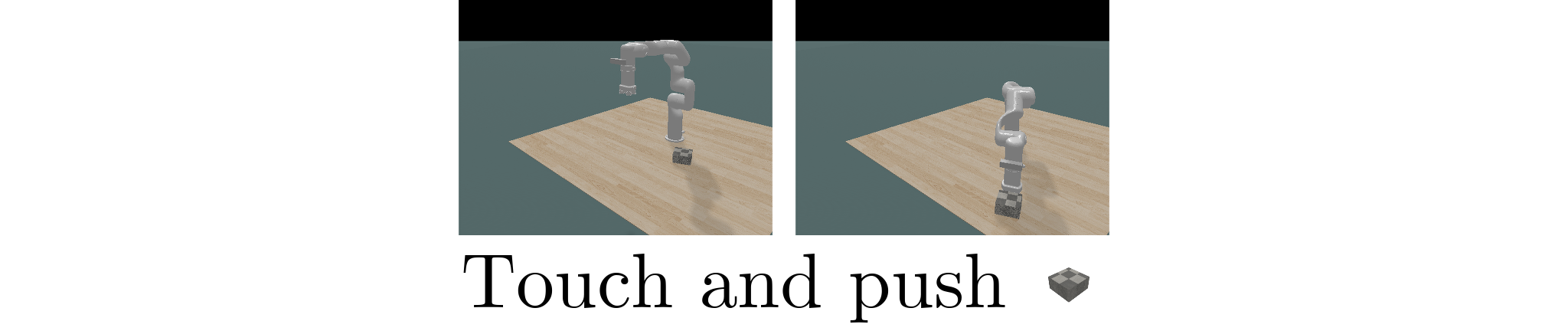}
    \end{center}
    \begin{itemize}
        \item \textbf{Prompts:} 
        \begin{enumerate}
            \item Touch and push \{obj:object\}$_{1}$.
        \end{enumerate}
        \item \textbf{Description:} Similar to 9, but now the specified object must move from its original position without the agent ever grasping it or the object toppling over.
        \item \textbf{Success Criteria:} The specified object must move at least $10$cm from its original position without it being grasped and the object should also not topple over.
    \end{itemize}

    \item \textbf{Touch topple:} 
    \begin{center}
        \includegraphics[width=0.9\linewidth]{figures/task_suite/Topple.png}
    \end{center}
    \begin{itemize}
        \item \textbf{Prompts:} 
        \begin{enumerate}
            \item Touch and topple \{obj:object\}$_{1}$, \{obj:object\}$_{2}$.
        \end{enumerate}
        \item \textbf{Description:} Similar to 9, but now the object must topple over. 
        \item \textbf{Success Criteria:} The specified object should be touched and toppled over i.e. there should be at least $45$ degree change in the vertical axis of the object.
    \end{itemize}

    \item \textbf{Trace:}
    \begin{center}
        \includegraphics[width=0.9\linewidth]{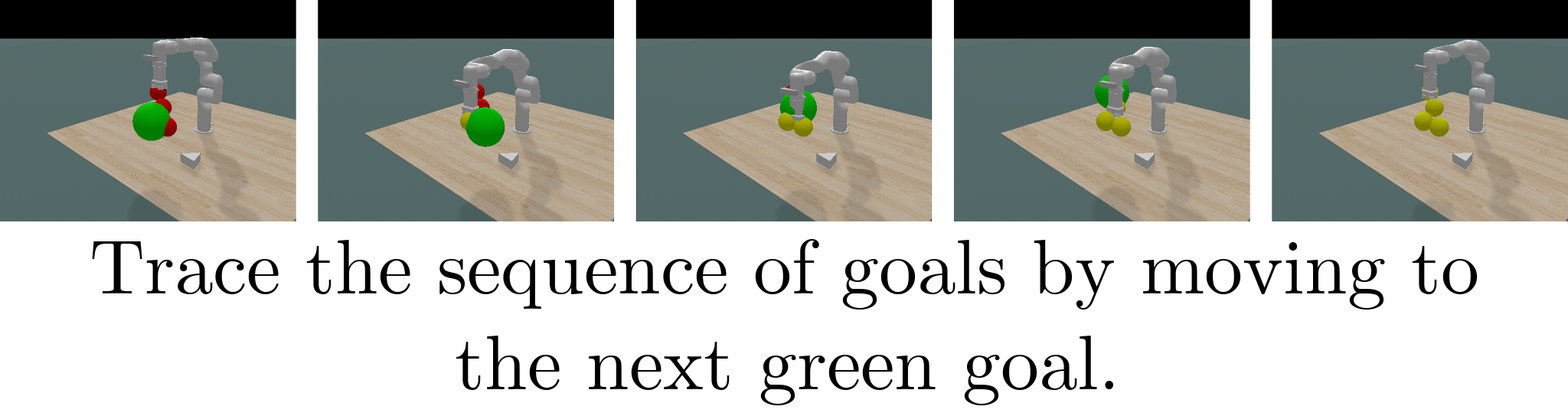}
    \end{center}
    \begin{itemize}
        \item \textbf{Prompts:} 
        \begin{enumerate}
            \item Trace the sequence of goals by moving to the next green goal.
        \end{enumerate}
        \item \textbf{Description:} The agent is tasked to touch goal positions specified by green spheres suspended in air. Once a goal is touched, it turns to yellow and another goal turns green. The agent can only succeed if it touches all the goals in the order of appearance. The number of goals may vary from $2$ to $5$.
        \item \textbf{Success Criteria:} All the goals are traced in order.

    \end{itemize}
    
\end{enumerate}

\subsection{L1: Intermediate Tasks}
The tasks included in L1 consists of simple compositions of skills acquired from L0 tasks.

\begin{enumerate}
    \item \textbf{Simple manipulation:} 
    \begin{center}
        \includegraphics[width=0.9\linewidth]{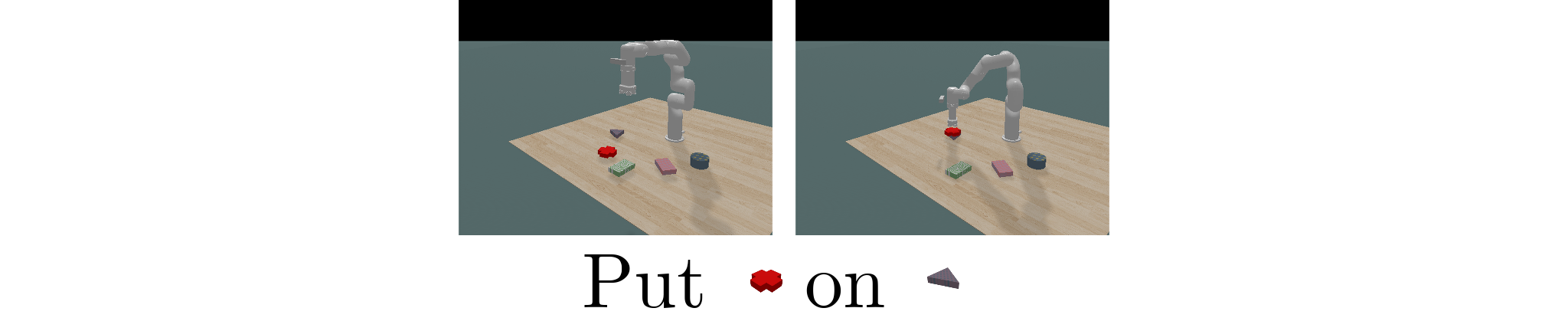}
    \end{center}
    \begin{itemize}
        \item \textbf{Prompts:} 
        \begin{enumerate}
            \item Put \{obj:object\}$_1$ on \{obj:object\}$_1$.
            \item Put object with \{tex:object\}$_1$ texture on object with \{tex:object\}$_2$ texture.
        \end{enumerate}
        \item \textbf{Description:} The tasks combined pick and place skills. The agent is tasked to pick a specified object \{obj:object\}$_1$ and put it on a goal object \{obj:object\}$_2$. In the final state, the target object must not be touch the end-effector or the ground. 
        \item \textbf{Success Criteria:} \{obj:object\}$_1$ is placed on top of \{obj:object\}$_2$ and the end-effector is not touching it. 
    \end{itemize}
    
    \item \textbf{Follow order:} 
    \begin{center}
        \includegraphics[width=0.9\linewidth]{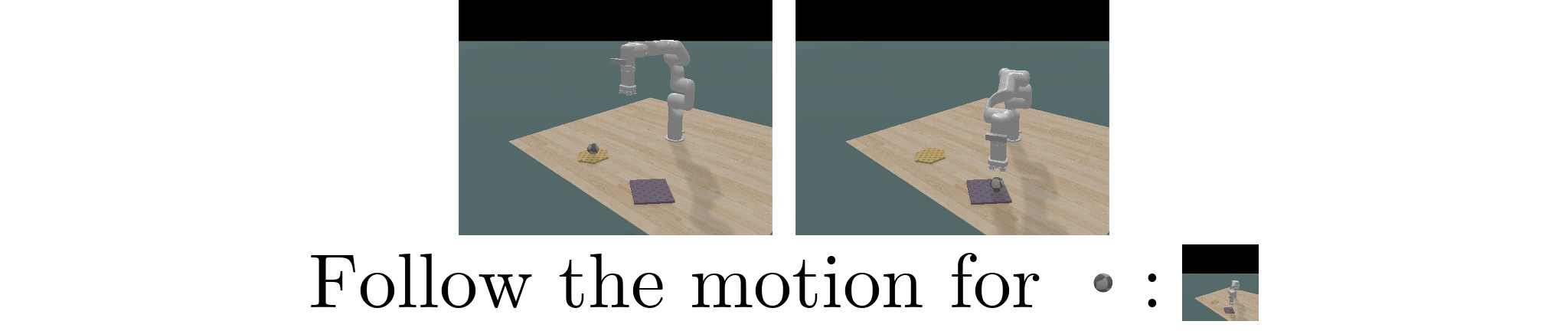}
    \end{center}
    \begin{itemize}
        \item \textbf{Prompts:} 
        \begin{enumerate}
            \item Follow the motion for \{obj:object\}$_1$: \{ks:keystep$_1$\}.
        \end{enumerate}
        \item \textbf{Description:} Given a specified object \{obj:object\}$_1$ and a set of goal states \{ks:keystep$_1$\}, the agent is tasked to achieve the goal states for the specified object in order. There may be multiple goal states which must be achieved in order.
        \item \textbf{Success Criteria:} All the goal states are achieved in order for the specified object.
    \end{itemize}

    \item \textbf{Follow order and restore:}
    \begin{center}
        \includegraphics[width=0.9\linewidth]{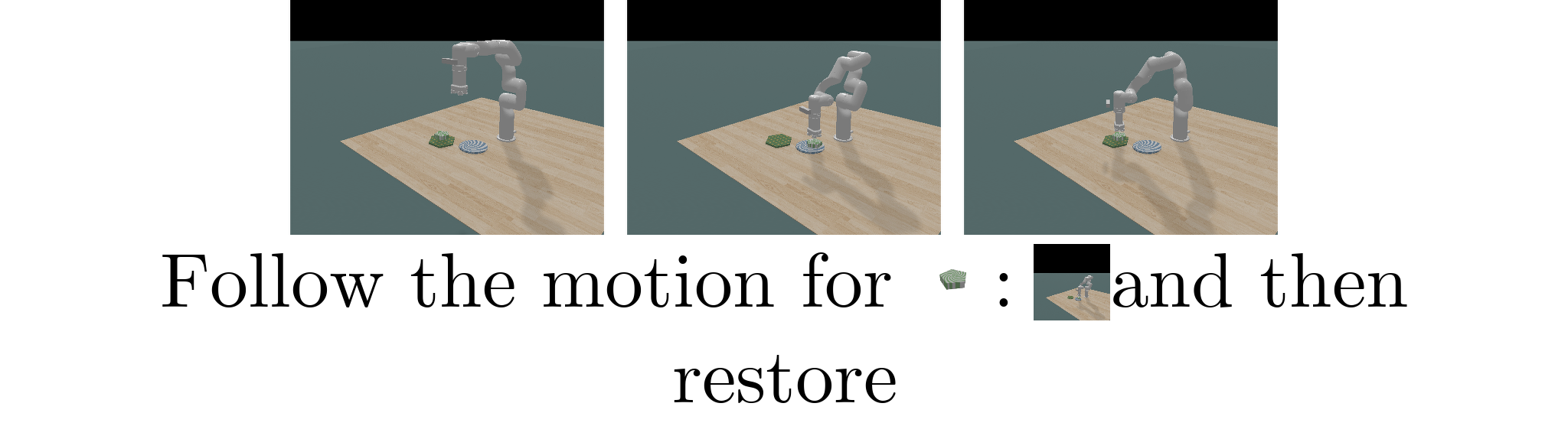}
    \end{center}
    \begin{itemize}
        \item \textbf{Prompts:} 
        \begin{enumerate}
            \item Follow the motion for \{obj:object\}$_1$: \{ks:keystep$_1$\} and then restore.
        \end{enumerate}
        \item \textbf{Description:} Similar to Task 2 in L1, with the additional constraint that the specified object must be returned to it's original state.
        \item \textbf{Success Criteria:} All the goal states are achieved in order for the specified object and then returned to it's initial state.
    \end{itemize}

    \item \textbf{Neighbour:} 
    \begin{center}
        \includegraphics[width=0.9\linewidth]{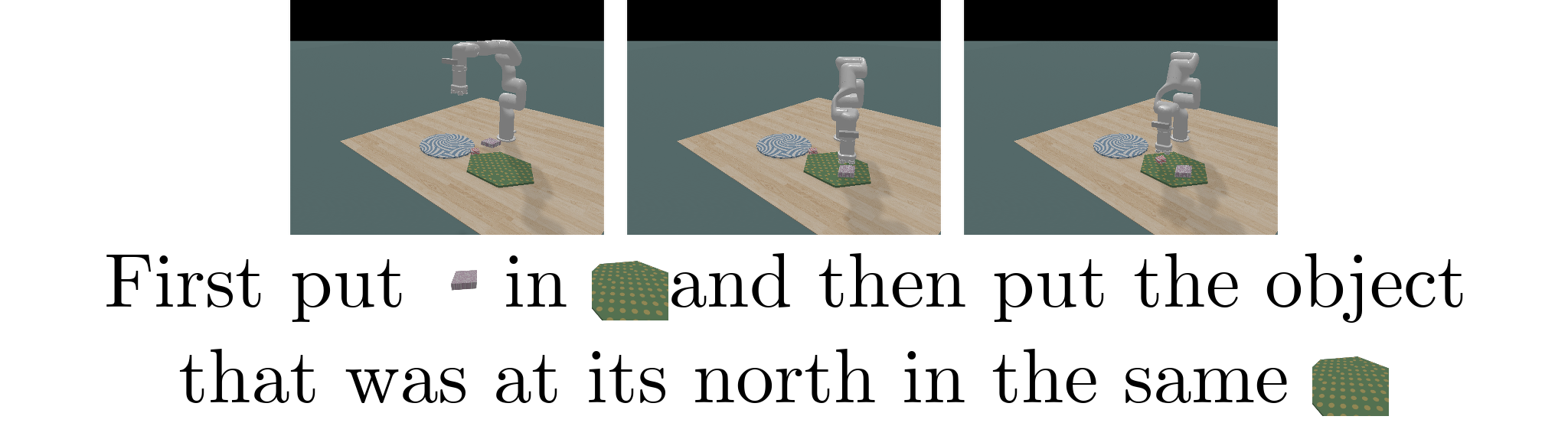}
    \end{center}
    \begin{itemize}
        \item \textbf{Prompts:} 
        \begin{enumerate}
            \item First put \{obj:object\}$_1$ in \{obj:object\}$_2$ and then put the object that was at its \{direction\} in the same \{obj:object\}$_2$
        \end{enumerate}
        \item \textbf{Description:} The agent is tasked to pick and place \{obj:object\}$_1$ in \{obj:object\}$_2$ and then pick and place the neighbour of \{obj:object\}$_1$ which was at a specific \{direction\} of \{obj:object\}$_1$. The direction can take values of north, south, east and west.
        \item \textbf{Success Criteria:} \{obj:object\}$_1$ and the specified neighbour, both be place in \{obj:object\}$_2$.
    \end{itemize}

    \item \textbf{Novel Adjective:} 
    \begin{center}
        \includegraphics[width=0.9\linewidth]{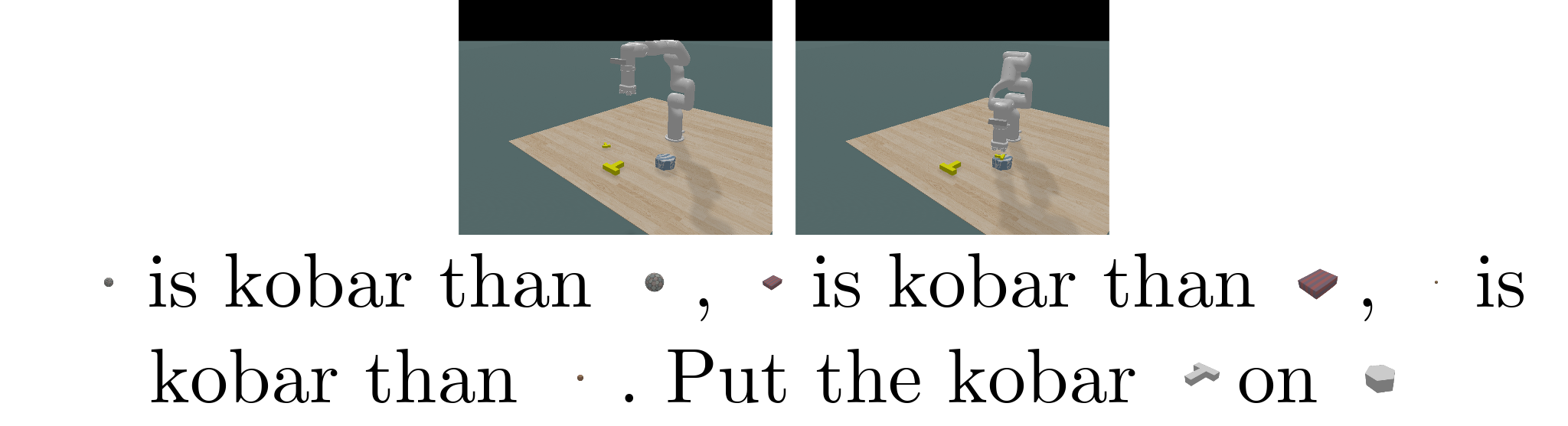}
    \end{center}
    \begin{itemize}
        \item \textbf{Prompts:} 
        \begin{enumerate}
            \item  \{obj:object\}$_1$ is \{adjective\} than  \{obj:object\}$_2$. Put the \{adjective\}  \{obj:object\}$_3$ on  \{obj:object\}$_4$.
        \end{enumerate}
        \item \textbf{Description:} The task is similar to Task 1 in L1, however instead of directly specifying the object with an image, the object is specified by an \{adjective\}. The \{adjective\} is a dummy adjective whose definition is conveyed by examples. The \{adjective\} can take values \textit{``daxer"}, \textit{``blicker"}, \textit{``modier"}, and \textit{``kobar"} which is chosen at random. For example, of ``kobar'' is chosen as an adjective which is supposed to mean taller, we initialize two meshes of the same object with different sizes where \{obj:object\}$_1$ is taller than \{obj:object\}$_2$. 

        \item \textbf{Success Criteria:} The target object corresponding to adjective is placed on the goal object. 

    \end{itemize}

    \item \textbf{Novel Noun:} 
    \begin{center}
        \includegraphics[width=0.9\linewidth]{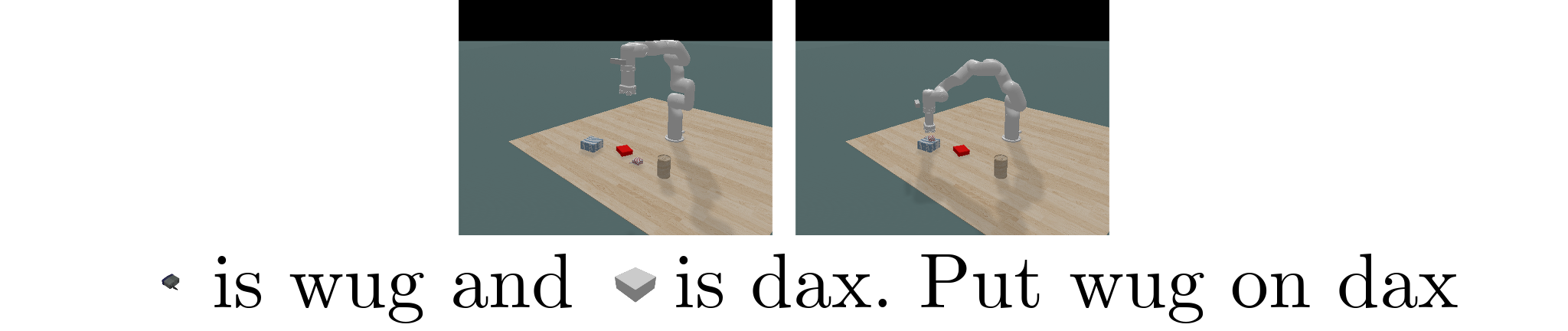}
    \end{center}
    \begin{itemize}
        \item \textbf{Prompts:} 
        \begin{enumerate}
            \item \{obj:object\}$_1$ is \{noun\}$_1$ and \{obj:object\}$_2$ is \{noun\}$_2$. Put \{noun\}$_1$ on \{noun\}$_2$.
        \end{enumerate}
        \item \textbf{Description:} This is similar to Task 5 in L1, however instead of an adjective the object is specified by a random noun. The \{noun\} can take values \textit{``dax"}, \textit{``blicket"}, \textit{``wug"} and \textit{``zup"} which is chosen at random. 

        \item \textbf{Success Criteria:} The correct target object is placed on the goal object.
    \end{itemize}

    \item \textbf{Novel Adjective and Noun:} 
    \begin{center}
        \includegraphics[width=0.9\linewidth]{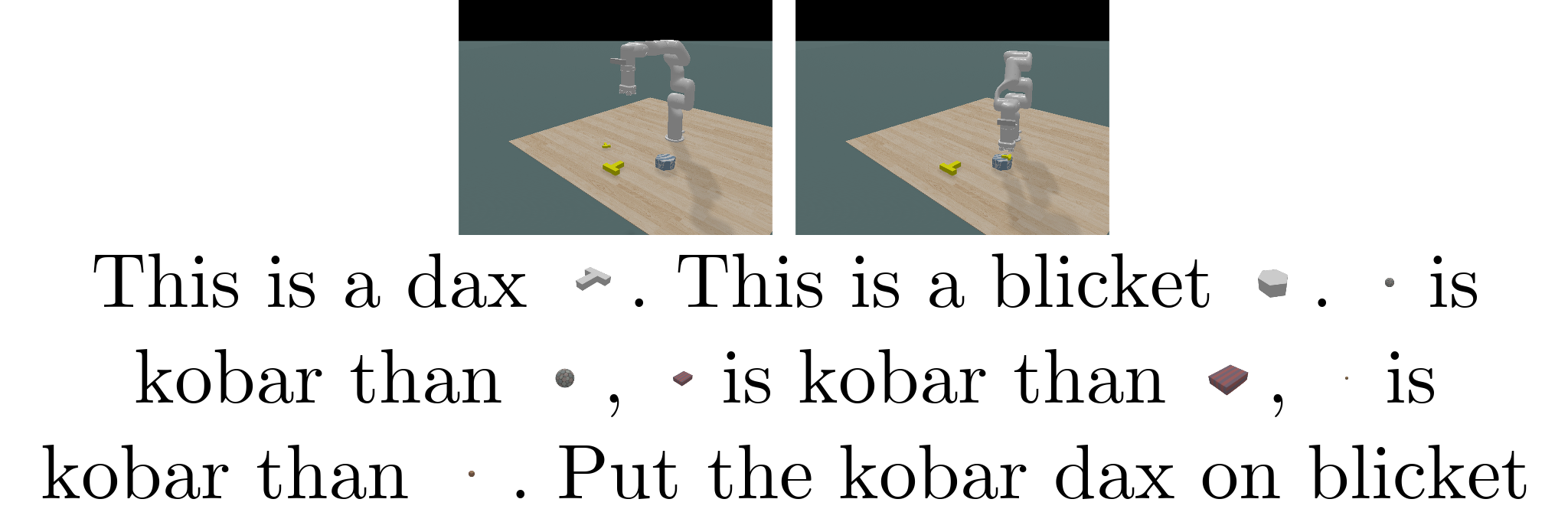}
    \end{center}
    \begin{itemize}
        \item \textbf{Prompts:} 
        \begin{enumerate}
            \item This is a \{noun\}$_1$ \{obj:object\}$_1$. This is a \{noun\}$_2$ \{obj:object\}$_2$. \{obj:object\}$_3$ is \{adjective\} than \{obj:object\}$_4$, \{obj:object\}$_5$ is \{adjective\} than \{obj:object\}$_6$. Put the \{adjective\} \{noun\}$_1$ on \{noun\}$_2$
        \end{enumerate}
        \item \textbf{Description:} This task is the combination of both Task 5 and Task 6 in L1. Here, the object specification involve both novel adjective and novel noun. 

        \item \textbf{Success Criteria:} The correct target object is placed on the goal object.
    \end{itemize}

    \item \textbf{Rearrange:} 
    \begin{center}
        \includegraphics[width=0.9\linewidth]{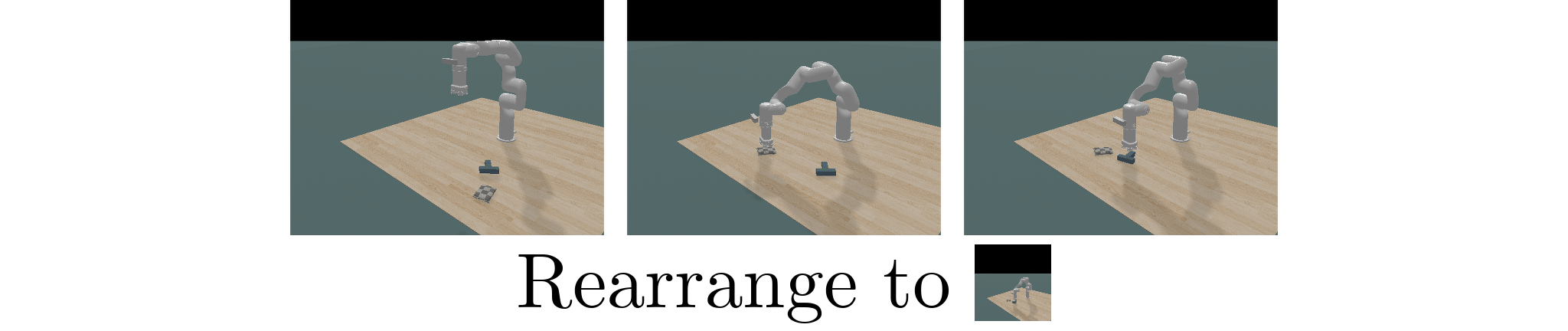}
    \end{center}
    \begin{itemize}
        \item \textbf{Prompts:} 
        \begin{enumerate}
            \item Rearrange to \{ks:scene\}
        \end{enumerate}
        \item \textbf{Description:} Here, the agent is tasked with rearranging the scene to object configuration shown in the scene image \{ks:scene\}.

        \item \textbf{Success Criteria:} All the objects in the scene are placed at the positions specified in the scene image \{ks:scene\}.
    \end{itemize}

    \item \textbf{Rearrange and restore:} 
    \begin{center}
        \includegraphics[width=0.9\linewidth]{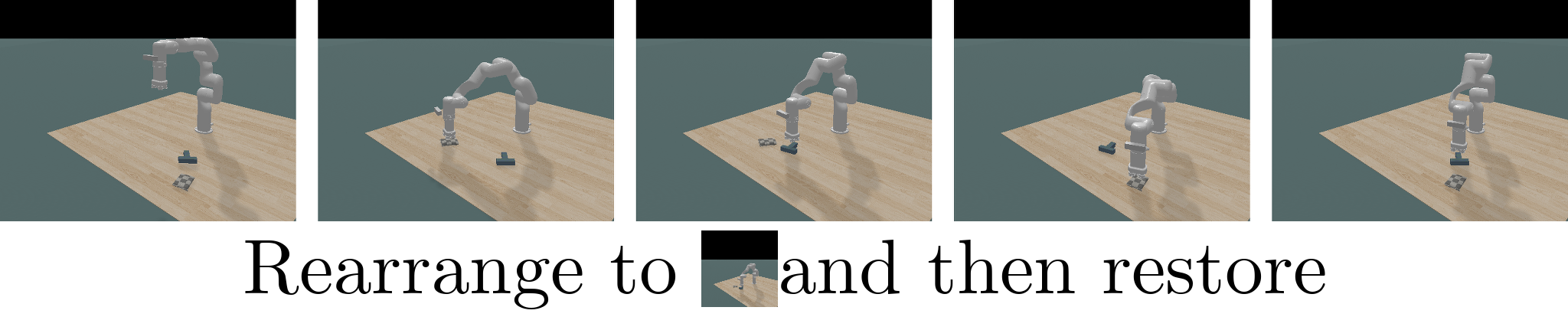}
    \end{center}
    \begin{itemize}
        \item \textbf{Prompts:} 
        \begin{enumerate}
            \item Rearrange to \{ks:scene\} and then restore.
        \end{enumerate}
        \item \textbf{Description:} This is similar to Task 8 in L1 with an additional constraint that after rearranging to the specified scene, the agent must bring the objects to their initial positions i.e. rearrange it back to the starting point. Note that the agent needs to remember where everything goes to be able to solve this.

        \item \textbf{Success Criteria:} All the objects in the scene are placed at the positions specified in the scene image \{ks:scene\} and once the rearrangement is complete they are brought back to the initial state.
    \end{itemize}

    \item \textbf{Rotate and restore:} 
    \begin{center}
        \includegraphics[width=0.9\linewidth]{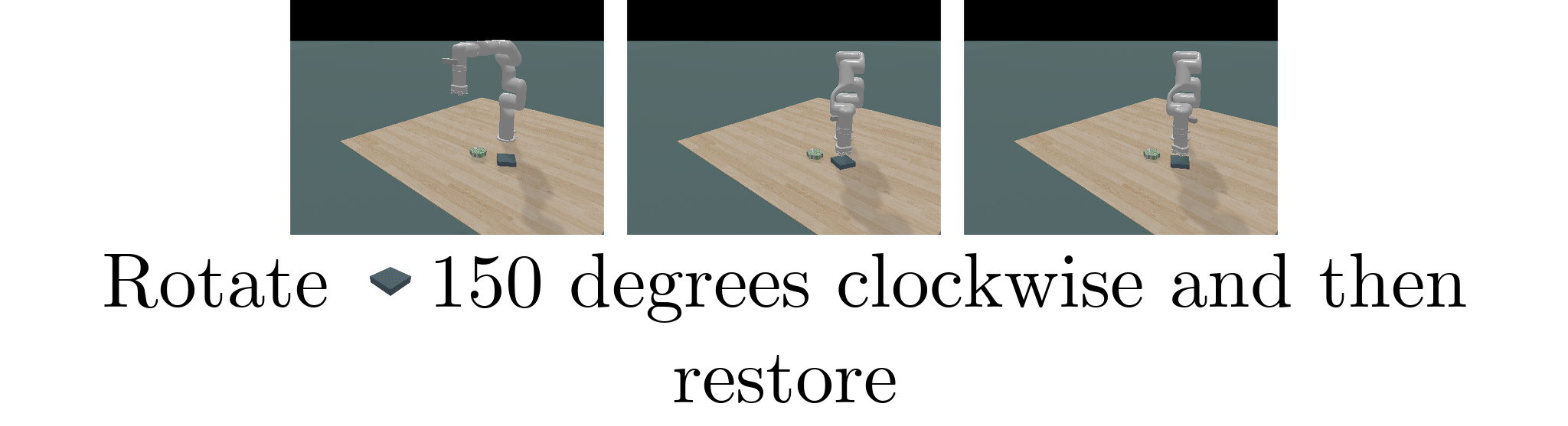}
    \end{center}
    \begin{itemize}
        \item \textbf{Prompts:} 
        \begin{enumerate}
            \item Rotate \{obj:object\} \{angle\} degrees \{direction\} and then restore
        \end{enumerate}
        \item \textbf{Description:} This task is similar to Task 6 in L0 with an additional constraint that once the rotation is complete, the agent needs to restore the object to its starting position.

        \item \textbf{Success Criteria:} The specified object is rotated in the correct direction within $5$ degrees of the specified angle. The position of the object should not change more than $5$cm and then restore the object to the starting point with the same criteria.
    \end{itemize}

    \item \textbf{Rotate symmetry:} 
    \begin{center}
        \includegraphics[width=0.9\linewidth]{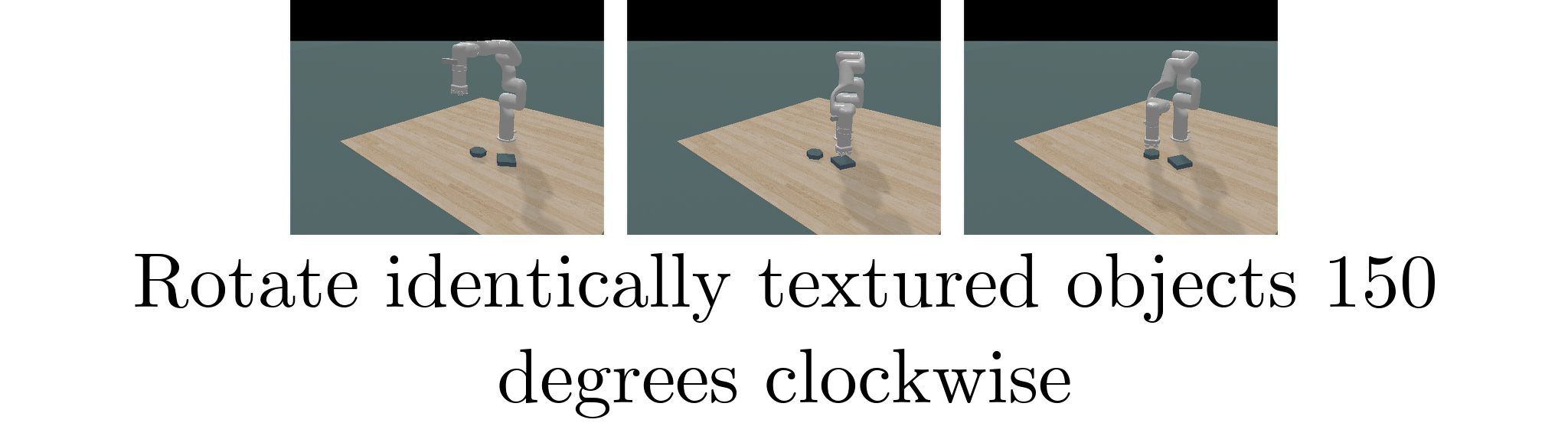}
    \end{center}
    \begin{itemize}
        \item \textbf{Prompts:} 
        \begin{enumerate}
            \item Rotate objects with \{tex:object\} texture \{angle\} degrees \{direction\}
            \item Rotate identically textured objects \{angle\} degrees \{direction\}
        \end{enumerate}
        \item \textbf{Description:} The task is similar to Task 6 in L0, however the object is specified using texture so the agent needs to select the correct object(s) among many distractor objects and rotate them by \{angle\} degrees in the correct \{direction\}.

        \item \textbf{Success Criteria:} All the objects with the specified texture are rotated by \{angle\} degrees in the correct \{direction\}.
    \end{itemize}

    \item \textbf{Stack:} 
    \begin{center}
        \includegraphics[width=0.9\linewidth]{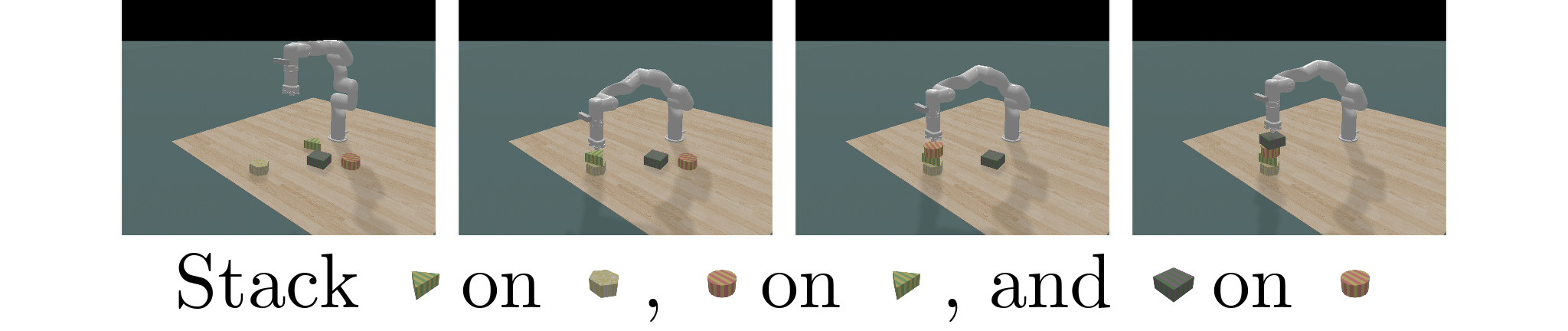}
    \end{center}
    \begin{itemize}
        \item \textbf{Prompts:} 
        \begin{enumerate}
            \item Stack \{obj:object\}$_1$ on \{obj:object\}$_2$, and \{obj:object\}$_3$ on \{obj:object\}$_1$
            \item Stack  object with \{tex:object\}$_1$ texture on object with \{tex:object\}$_2$  texture, object with \{tex:object\}$_3$ texture on object with \{tex:object\}$_1$  texture
            \item Stack objects as in \{ks:keystep\}$_2$
            \item Stack objects in this order \{ks:keystep\}$_0$ \{ks:keystep\}$_1$ \{ks:keystep\}$_2$
        \end{enumerate}
        \item \textbf{Description:} The agent is tasked to stack multiple objects on top of each other in the order as specified in the prompt. 

        \item \textbf{Success Criteria:} The objects in the scene are stacked in the correct order.
    \end{itemize}

    \item \textbf{Stack reversed:} 
    \begin{center}
        \includegraphics[width=0.9\linewidth]{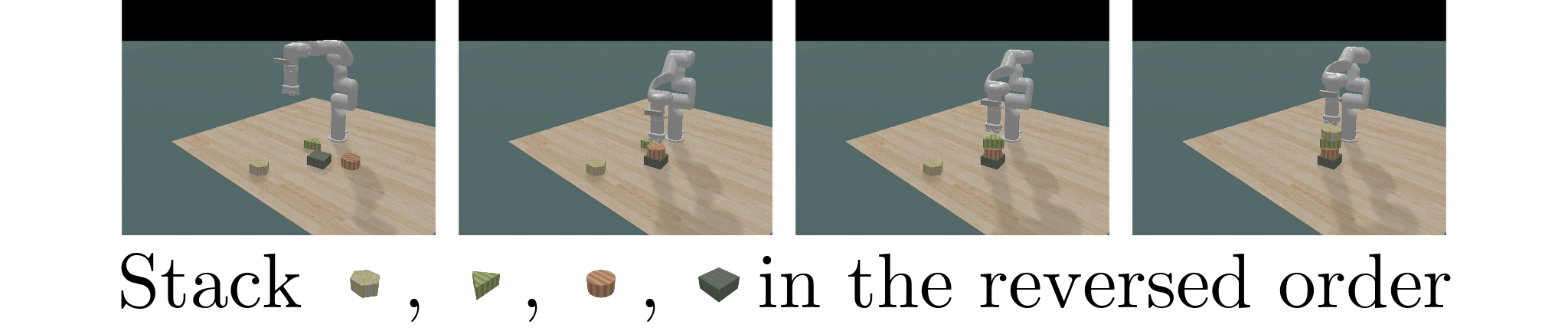}
    \end{center}
    \begin{itemize}
        \item \textbf{Prompts:} 
        \begin{enumerate}
            \item Stack  \{obj:object\}$_1$,  \{obj:object\}$_2$,  \{obj:object\}$_3$,  \{obj:object\}$_4$ in the reversed order
        \end{enumerate}
        \item \textbf{Description:} The task is similar to Task 12 in L1 however the agent is tasked to stack the specified objects in the reverse order.

        \item \textbf{Success Criteria:} The objects in the scene are stacked in the reverse order of what is specified.
    \end{itemize}

    \item \textbf{Sort:} 
    \begin{center}
        \includegraphics[width=0.9\linewidth]{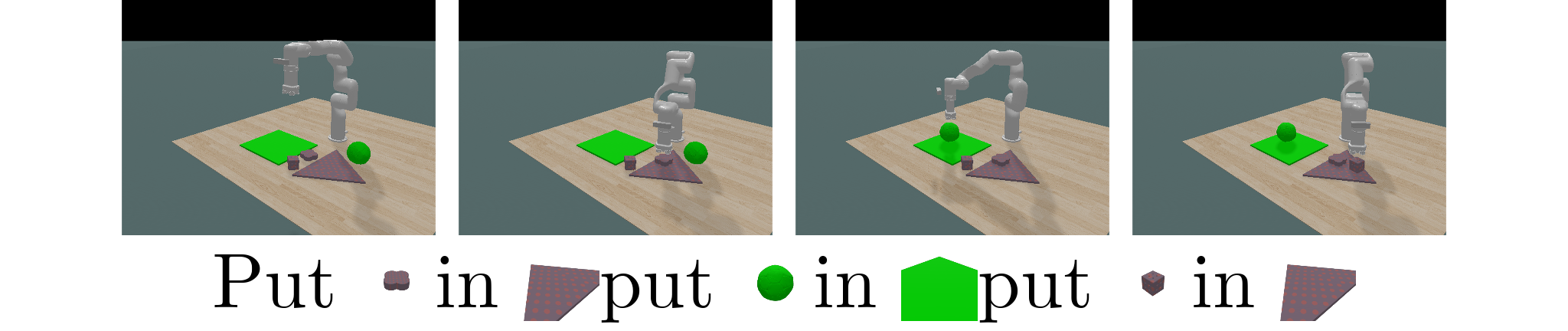}
    \end{center}
    \begin{itemize}
        \item \textbf{Prompts:} 
        \begin{enumerate}
            \item Place the objects in the identically textured areas
        \end{enumerate}
        \item \textbf{Description:} In this task, the agent is required to place objects in the areas with similar texture as the object.

        \item \textbf{Success Criteria:} All the objects are placed in the areas which have same texture as the objects.
    \end{itemize}

    \item \textbf{Swap:} 
    \begin{center}
        \includegraphics[width=0.9\linewidth]{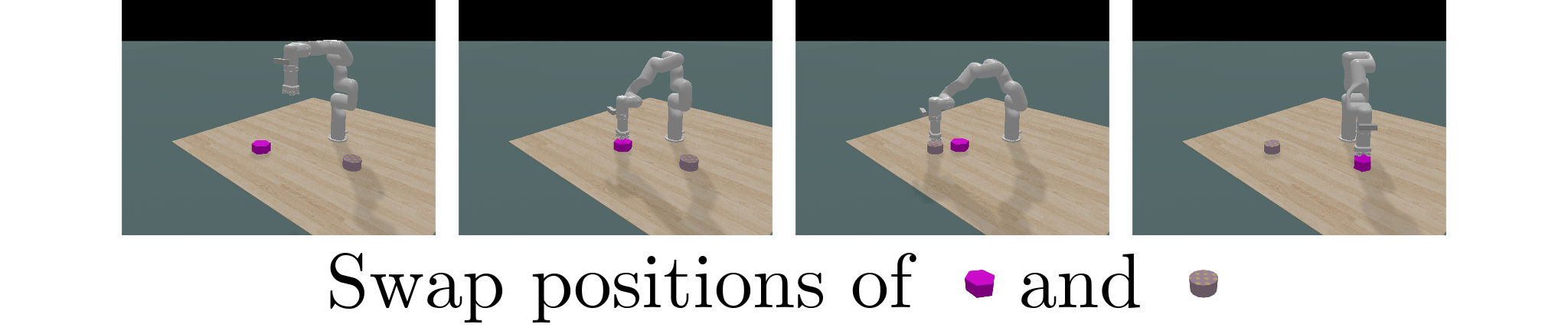}
    \end{center}
    \begin{itemize}
        \item \textbf{Prompts:} 
        \begin{enumerate}
            \item Swap positions of \{obj:object\}$_1$ and \{obj:object\}$_2$
        \end{enumerate}
        \item \textbf{Description:} The agent is tasked to swap the positions of two objects as specified in the prompt. The only way to achieve the result is to move one of the object away and then place the other object in its place and then repeating the same with the initial object.

        \item \textbf{Success Criteria:} The positions of the two objects are swapped.
    \end{itemize}
    
\end{enumerate}

\subsection{L2: Complex Tasks}
The tasks included in L2 consists of more complex compositions of skills acquired from L0 and L1 tasks.

\begin{enumerate}

    \item \textbf{Balance:} 
    \begin{center}
        \includegraphics[width=0.9\linewidth]{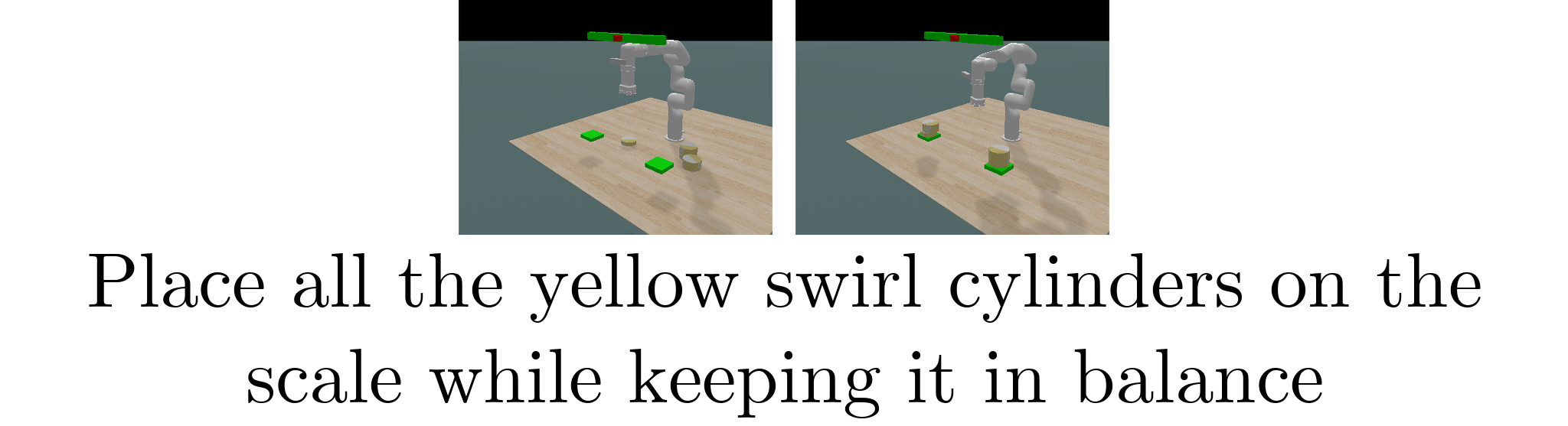}
    \end{center}
    \begin{itemize}
        \item \textbf{Prompts:} 
        \begin{enumerate}
            \item Place all the objects on the scale while keeping it in balance
        \end{enumerate}
        \item \textbf{Description:} The agent is tasked to balance a weight scale by placing the objects with appropriate weights on either side. The weight of the objects is proportional to their size. We initialize objects in a way that there is always a split of two sets of objects which add up to the same weight. 

        \item \textbf{Success Criteria:} All the objects are placed on the scale and it is balanced.
    \end{itemize}

    \item \textbf{Sort Stack:} 
    \begin{center}
        \includegraphics[width=0.9\linewidth]{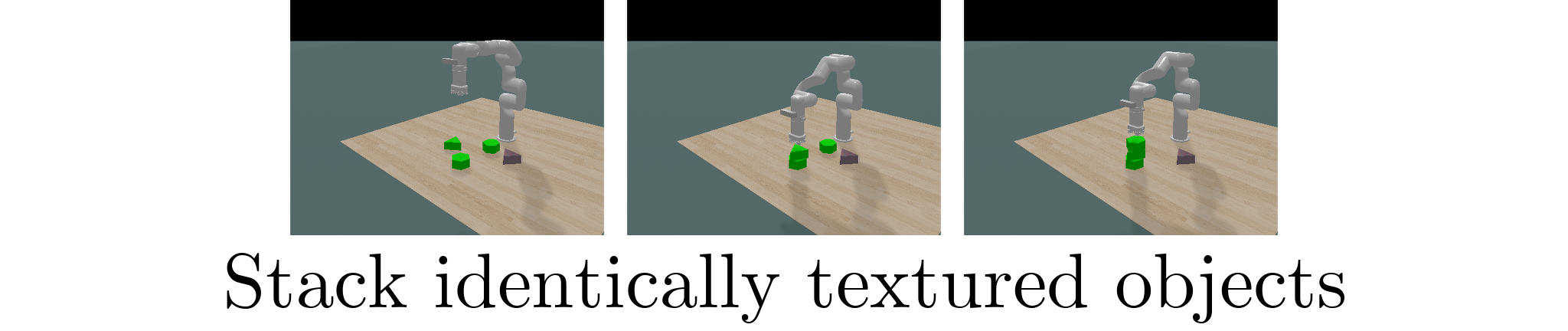}
    \end{center}
    \begin{itemize}
        \item \textbf{Prompts:} 
        \begin{enumerate}
            \item Stack identically textured objects
            \item Place identically textured objects on top of each other
        \end{enumerate}
        \item \textbf{Description:} The task is a composition of Task 12 and Task 14 in L1. The agent here is required to sort the objects according to their texture however, instead of just placing the objects on an area, it is tasked to stack them on top of each other.

        \item \textbf{Success Criteria:} All the objects with identical textures are stacked on top of each other.

    \end{itemize}

    \item \textbf{Stack topple:} 
    \begin{center}
        \includegraphics[width=0.9\linewidth]{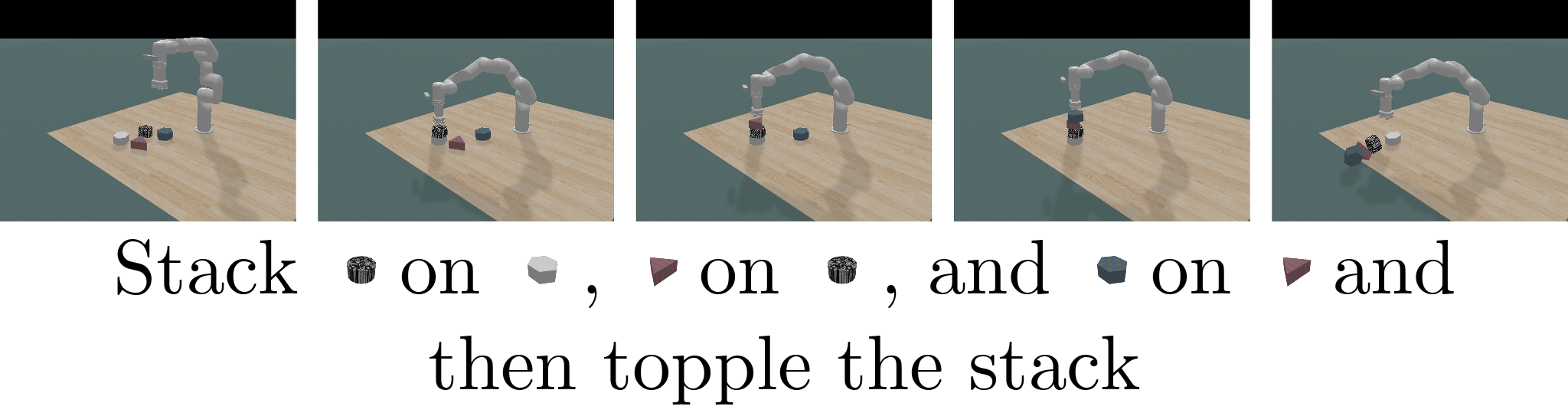}
    \end{center}
    \begin{itemize}
        \item \textbf{Prompts:} 
        \begin{enumerate}
            \item Stack \{obj:object\}$_1$ on \{obj:object\}$_2$, and \{obj:object\}$_3$ on \{obj:object\}$_1$ and then topple the stack
            \item Stack  object with \{tex:object\}$_1$ texture on object with \{tex:object\}$_2$  texture, object with \{tex:object\}$_3$ texture on object with \{tex:object\}$_1$  texture and then topple the stack
            \item Stack objects as in \{ks:keystep\}$_2$ and then topple the stack
            \item Stack objects in this order \{ks:keystep\}$_0$ \{ks:keystep\}$_1$ \{ks:keystep\}$_2$ and then topple the stack
        \end{enumerate}
        \item \textbf{Description:} The task is a compostion of Task 11 in L0 and Task 12 in L1 where the agent is tasked to first stack the objects as specified in the prompt and then topple the resulting stack.

        \item \textbf{Success Criteria:} The objects are first stacked as specified in the prompt and then are toppled such that all the objects end up on the ground.
    \end{itemize}

    \item \textbf{Swap with push:} 
    \begin{center}
        \includegraphics[width=0.9\linewidth]{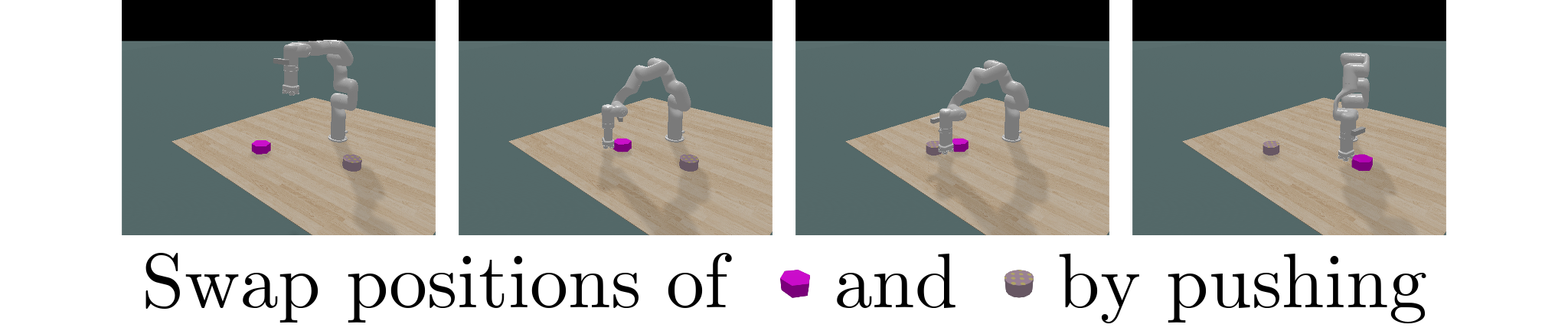}
    \end{center}
    \begin{itemize}
        \item \textbf{Prompts:} 
        \begin{enumerate}
            \item Swap positions of \{obj:object\}$_1$ and \{obj:object\}$_2$ by pushing
        \end{enumerate}
        \item \textbf{Description:} The task is similar to Task 15 in L1 but instead of swapping by pick and place skills, the agent is tasked to do the same by pushing the objects.

        \item \textbf{Success Criteria:} The positions of the two objects are swapped without the objects being grasped.
    \end{itemize}

    \item \textbf{Swap and rotate:} 
    \begin{center}
        \includegraphics[width=0.9\linewidth]{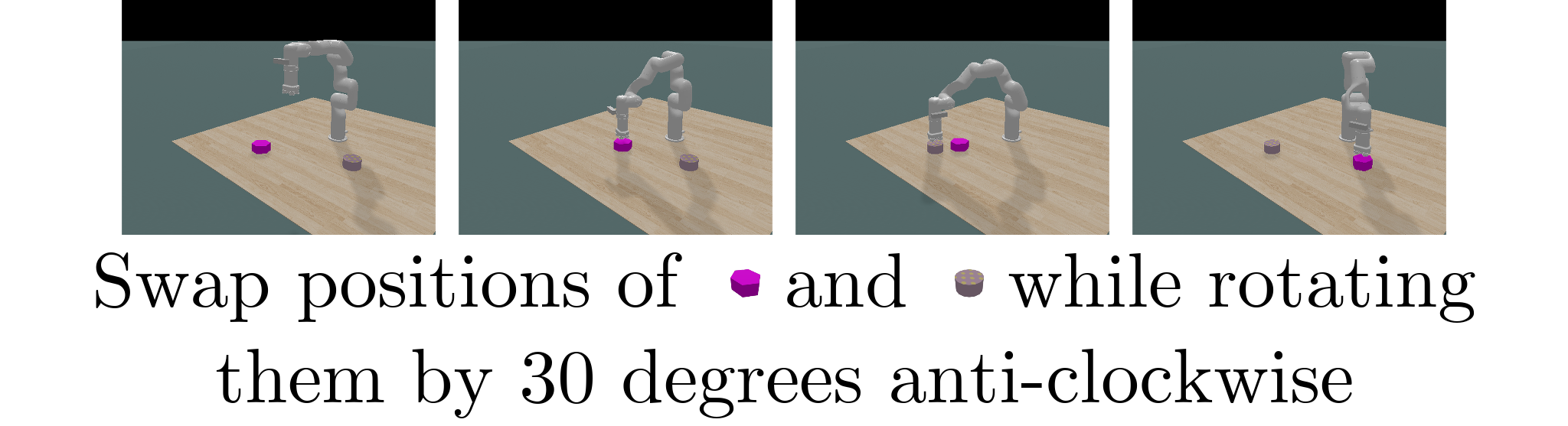}
    \end{center}
    \begin{itemize}
        \item \textbf{Prompts:} 
        \begin{enumerate}
            \item Swap positions of \{obj:object\}$_1$ and \{obj:object\}$_2$ but rotate them by \{angle\} degrees \{direction\}
        \end{enumerate}
        \item \textbf{Description:} The task is similar to Task 15 in L1 with an additional constraint that the objects must be rotated by \{angles\} degrees in \{direction\}.

        \item \textbf{Success Criteria:} The positions of the two objects are swapped and the objects are rotated by \{angles\} degrees in \{direction\} with respect to their initial pose.
    \end{itemize}

    \item \textbf{Throw sort (sort by throwing):} 
    \begin{center}
        \includegraphics[width=0.9\linewidth]{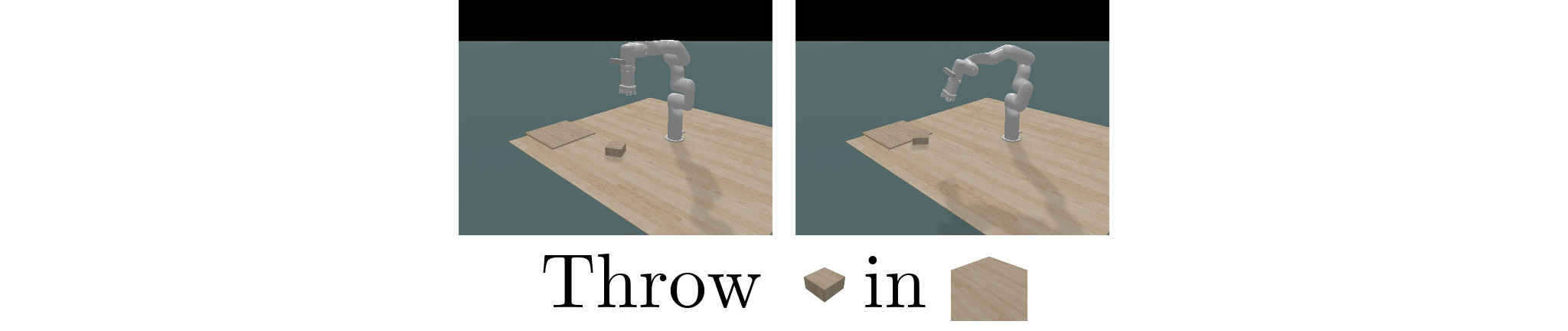}
    \end{center}
    \begin{itemize}
        \item \textbf{Prompts:} 
        \begin{enumerate}
            \item Place the objects in the identically textured areas by throwing
        \end{enumerate}
        \item \textbf{Description:} The task is similar to Task 14 in L1 but instead of sorting by picking and placing, the agent needs to throw the objects into specified areas. To force the robot to using throwing skill, the areas are position out of reach of the robot such that the task is impossible to complete without throwing.

        \item \textbf{Success Criteria:} All the objects are placed in the areas which have same texture as the objects and the objects are ``thrown'' in the areas instead of ``placed''.
    \end{itemize}
    
\end{enumerate}

\section{ClevrSkills dataset}
ClevrSkills includes 330k robot episodes/trajectories including videos (from multiple views), corresponding actions, and other annotations including text, bounding boxes, camera poses, etc., which were generated from over $33$ tasks in the ClevrSkills environment suite. It includes a carefully designed curriculum of tasks which can be used for training robotics models to perform tasks ranging from simple pick and place to more complicated manipulation tasks, such as sorting, stacking etc.

\subsection{Dataset structure}
The dataset consists of $33$ zip files, each containing data for one task. %
The archive contains directories named ``traj\_\{seed\}'' where seed denotes the seed used to generate the episode in the ClevrSkills environments 

Each directory corresponds to one episode which contains, 
\begin{itemize}
    \item Videos (from multiple cameras – these are used as inputs for models)
    \item Actions (Nx7 matrix in .npy format where N denotes the length of the episode) 
    \item Action labels (.npy file including text label for each action step)
    \item Camera parameters (.npy files for camera parameters of each camera)
    \item ep\_info.json (meta data of the episode)
    \item info.json (task specification / prompts and textures used in the episode)
    \item Keysteps (images of keysteps of the task)
    \item prompt\_assets.npy (images used in the prompts) 
    \item rewards.npy (reward for each step in the episode) 
    \item succes.npy (a label denoting if the task was successful or not at each timestep).
\end{itemize}
 
We also include a detailed datasheet in Appendix.~\ref{sec:datasheet}. The files can be downloaded from \url{https://www.qualcomm.com/developer/software/clevrskills-dataset}.

\section{Baseline architectures}
\label{app:baselines}

\subsection{Jack of All Trades (JAT)}
We use the open-source JAT model which is a transformer model trained to handle both \textit{text-centric} and \textit{sequential decision making} tasks. We focus on sequential decision making tasks as ClevrSkills fall into this class of tasks. For such tasks, the episodes are pre-processed to produce sequences of observation and action embeddings denoted as $[(\phi(s_0), \phi(a_0)), (\phi(s_1), \phi(a_1)), ...]$ where $s_i$ is image observation and $a_i$ is corresponding action at step $i$, and $\phi$ is an input dependent embedding function. The embedding function used depends on the input as follows:
\begin{itemize}
    \item \textbf{Images observation:} The input image is resized to $84\times84$ and passed through three blocks each consisting of a convolutional layer, an instance normalization layer and an attention layer. The resulting features are flattened and passed through a linear layer to produce an embedding of size $768$.
    \item \textbf{Continuous actions:} The continuous action vector is padded to achieve a length of $377$, corresponding to the maximum achievable continuous observation in JAT dataset. The padded vector is passed through a linear layer to produce an embedding of size $768$.
    \item \textbf{Text data:} Text data is tokenized using the same strategy as GPT-2~\cite{radford2019language}.
\end{itemize}

Different from the JAT dataset, ClevrSkills tasks cannot differentiated based on the starting frame and therefore, require explicit task specification. Therefore, we modify the input sequences to include a multi-modal prompt at the start so that the sequence is modified as $[p_0, p_1, ..., p_n, (\phi(s_0), \phi(a_0)), (\phi(s_1), \phi(a_1)), ...]$, where $p_i$ denotes the tokens from the multi-modal prompt which could either be tokenized text or image embedding. Since the sequences in JAT only take one input image, we only make use of the ``base'' camera from ClevrSkills dataset as our image stream. 

We start with a trained JAT model as initialization and fine-tune the model on the ClevrSkills dataset with MSE loss.

\subsection{RoboFlamingo}
RoboFlamingo uses an off-the-shelf Flamingo based vision language model as the feature extractor for language instructions and the image input in robotics tasks. The resulting features from the model are then passed off to an LSTM based policy head to predict low-level continuous actions. Concretely, given a language instruction $L$ and sequence of image and action pairs $[(s_0, a_0), (s_1, s_1), ...]$ the input is processed into pairs of language instruction and images $[(L, s_0), (L, s_1), ...]$, which are passed through a Flamingo model to produce output embeddings $X_t = \{x_0, x_1, ..., x_t\}$. The output embeddings are then passed through a pooling layer to produce an embedding for the sequence $\tilde{X_t} = MaxPooling(X_t)$, which is then passed to the LSTM based policy head to predict the continuous action $a_t = LSTM(\tilde{X_t})$. Since ClevrSkills includes multi-modal prompts in place of language only prompts, we modify the input sequence as $[(L, s_{p0}), ..., (L, s_{pm}), (L, s_0), (L, s_1), ...]$ where the first $M$ language-image pairs correspond to the multi-modal prompt. Note that the language instruction $L$ is processed by the language model, and the images $s_i$ are processed by a Perceiver model which are used in cross attention layers in later layers of the LLM.
We use the best performing RoboFlamingo model from \cite{li2023vision}. 

\subsection{StreamRoboLM}
StreamRoboLM (Streaming Robotics Language Model) is based on an LRR like model~\cite{bhattacharyya2023look} %
that takes streaming video as input. 
This is different from RoboFlamingo as the base vision language model is only used to extract features from one image at a time, whereas StreamRoboLM can reason over the whole video at the same time. We follow the Flamingo~\cite{alayrac2022flamingo} model and use \textit{<image>} tokens which are used in cross attention layers to cross attend to image embeddings generated by a ViT~\cite{dosovitskiy2020image}. 
After multiple self/cross-attention layers, the model outputs embeddings for each input token. These embeddings are then fed to an LSTM policy head which predicts the low-level actions. Concretely, given a multi-modal prompt as task specification $L = [p_1, p_1, ..., p_m]$ where $p_i$ can either be a text token or an image, and sequence of state image and action pairs $[(s_1, a_1), (s_2, s_2), ...]$ the input is processed such that we replace the images in multi-modal prompt with an \textit{<image>} token, and append an \textit{<image>} token for each state image $s_i$ in the input sequence which results in a input sequence of $I = [p'_1, ..., p'_m, q_1, ..., q_n]$, where $p'_i$ denotes either a text token or \textit{<image>} token in the prompt and $q_i$ denotes an \textit{<image>} token for corresponding state image $s_i$. Given this input $I$ and the sequence of prompt and state images the model produces embeddings for each of the input token $[e_1, ..., e_m, e_{m+1}, ..., e_n]$ where the first $M$ embeddings correspond to the prompt. We input the non-prompt embeddings $[e_{m+1}, ..., e_n]$ into the LSTM, which in turn produces the low-level actions $[a'_{m+1}, ..., a'_n]$. We use MSE loss over predicted and ground truth actions to train the network.

\section{Training Details}
\label{sec:training}
\begin{table*}[t]
  \centering
  \scriptsize
  \begin{tabularx}{0.70\linewidth}{lcccc}
    \toprule
    \textbf{Hyperparams} & \textbf{JAT} & \textbf{RoboFlamingo} & \textbf{Octo} & \textbf{StreamRoboLM} \\
    \midrule
    Learning rate & 1e-4 & 1e-6 & 3e-5 & 1e-6 \\
    Optimizer & AdamW & AdamW & AdamW & AdamW \\
    Batch Size / GPU & 64 & 6 & 64 & 12 (opt), 4 (llama3) \\
    Input images & ``base'' & ``base'', ``hand'' & ``base'', ``hand'' & ``base'', ``hand'' \\
    Image resolution & $84\times84$ & $256\times256$ & $256\times256$ & $256\times256$ \\
    \bottomrule
  \end{tabularx}
  \caption{Hyper-parameters used in training.}
  \label{tab:hyper}
\end{table*}

For JAT, Octo and RoboFlamingo, we use the official open-source code bases to run the experiments on ClevrSkills data. We implement StreamRoboLM in Pytorch~\cite{paszke2019pytorch}. The main training hyper-parameters are shown in Table~\ref{tab:hyper}. Additionally since some of the tasks like ``rotate'' and ``swap'' require memory of the initial state, we train all the baselines with the first frame as part of the prompt i.e. we treat the image of the initial state of the environment as part of the task prompt. We train the baselines until convergence and evaluate every 50k iterations and report the best results.

\begin{figure*}[t!]
    \centering
    \begin{subfigure}[t]{0.5\textwidth}
        \centering
       \includegraphics[width=1.0\linewidth]{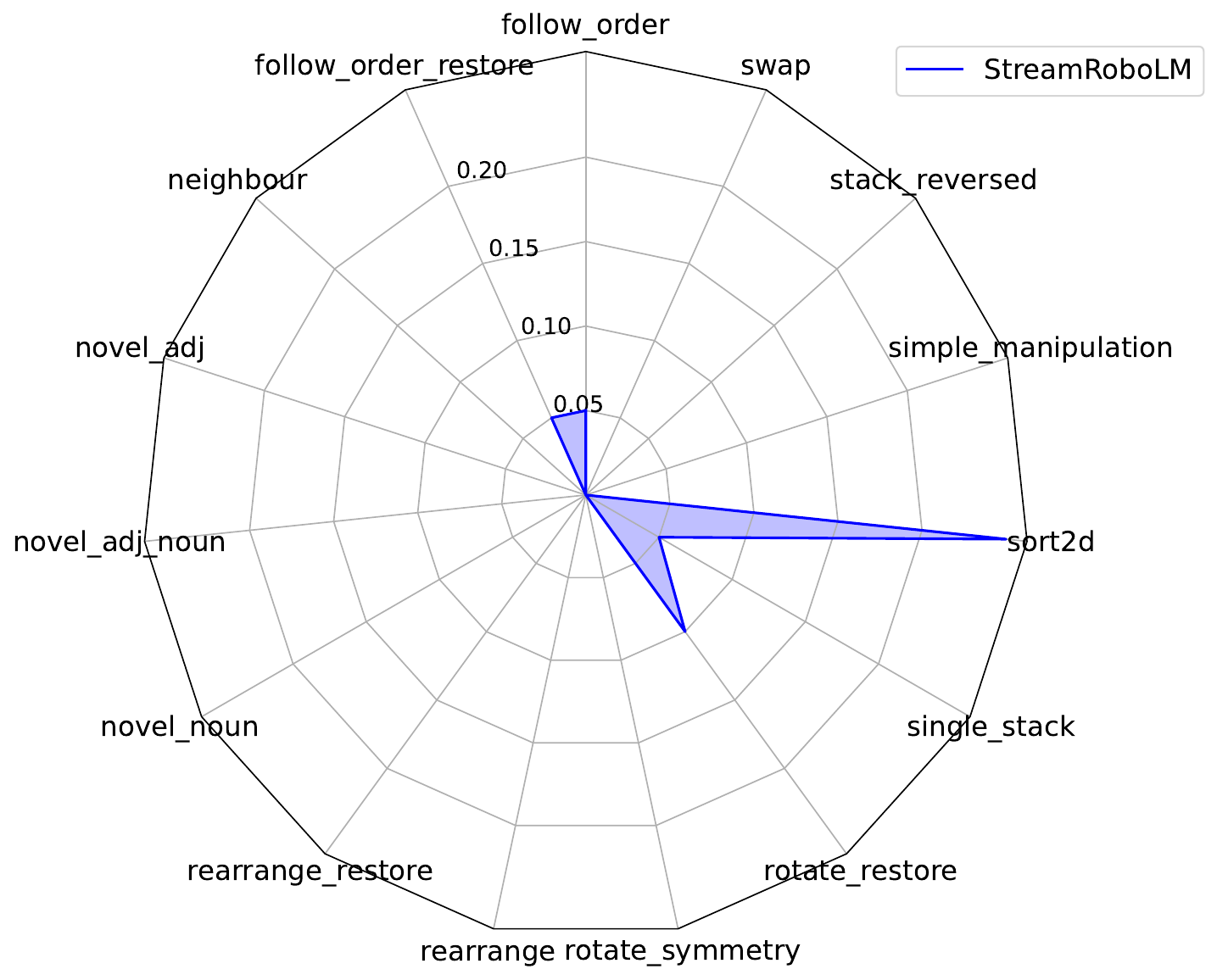}
    \end{subfigure}%
    ~ 
    \begin{subfigure}[t]{0.5\textwidth}
        \centering
       \includegraphics[width=1.0\linewidth]{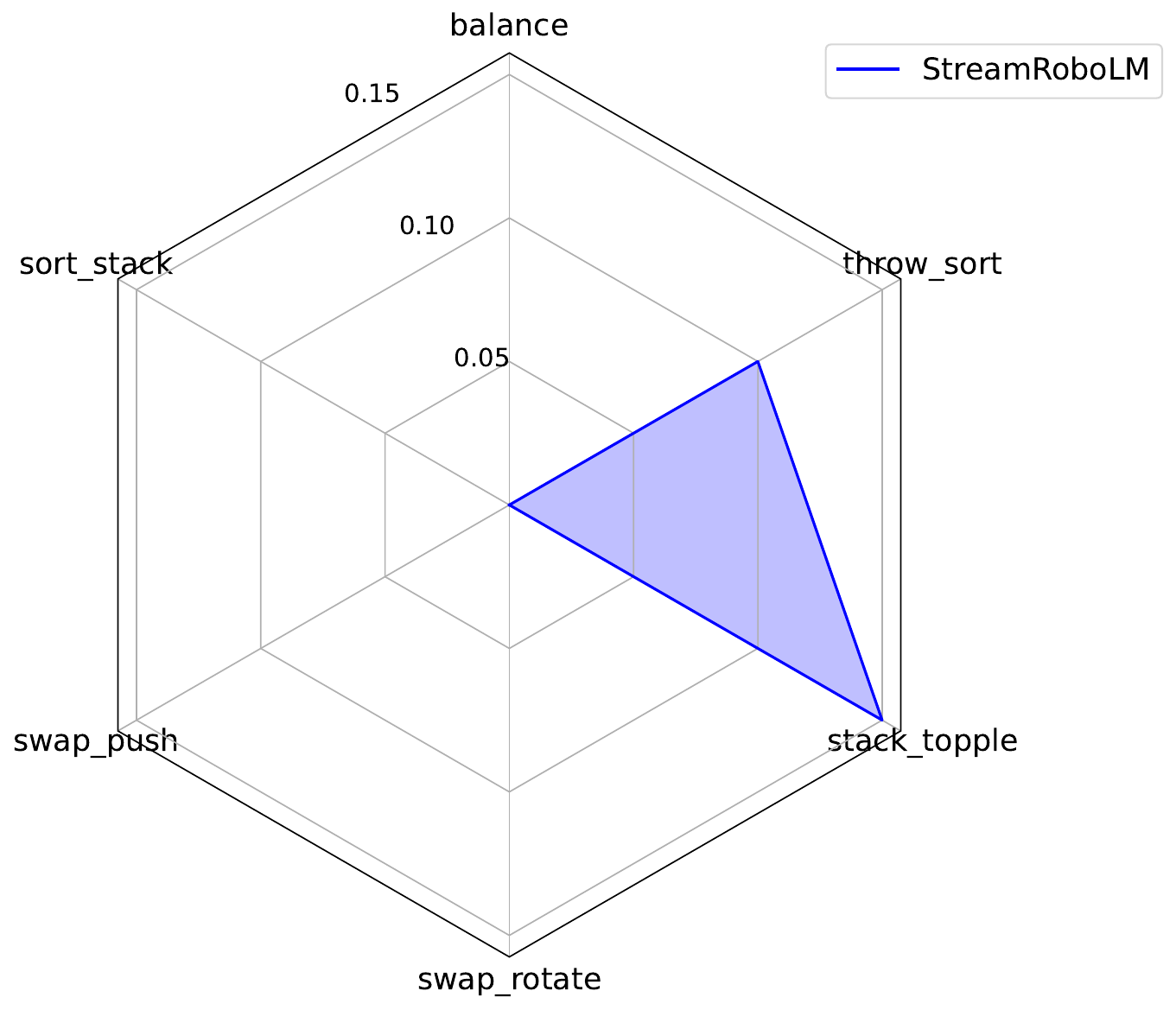}
    \end{subfigure}
    \caption{\emph{Left}: Task-wise success rate of StreamRoboLM (opt) on L1 tasks. \emph{Right}: Task-wise success rate of StreamRoboLM (opt) on L2 tasks.}
    \label{fig:task_wise_l1l2}
\end{figure*}

\rebuttal{\section{Task-wise performance on L1 and L2}
In Figure~\ref{fig:task_wise_l1l2}, we show the task-wise success rate of the StreamRoboLM (Opt) baseline on L1 and L2 tasks. We note that although, it struggles to get good performance overall, it achieves decent performance on some of the tasks. This shows that even within L1 and L2 tasks, some tasks are much harder than others. For example, StreamRoboLM (Opt) gets $25\%$ success rate on ``sort'' task which involve moving objects to identically textured areas. These areas can be large in size and therefore the policy does not require fine motor control for placement. ``Simple manipulation'', in comparison, is a much harder task as it first requires careful selection of the target top and base objects and then the policy also requires fine motor control to place the top object on a similarly sized base object. Tasks involving swapping and rearranging are also specially challenging because they not only require the policy to infer the right positions of objects from 2d images, they also require the policy to ``remember'' the original positions of the objects. Similarly in L2, ``stack and topple'' seem to be the easiest task as it is the simplest composition where the policy first needs to stack and then topple. Other tasks in L2 are much harder as the skills required are ``superimposed'' e.g. ``swap push'' not only requires swapping positions of the two objects, it also needs to be achieved by pushing instead of pick and place skills which makes it a specially hard task. On a high level, we note that inclusion of tasks with such weaker compositions allows for easier/better signal for progress towards solving compositional understanding in robotics.}

\section{Language-only prompt results}
\label{app:lang_only}
\begin{figure}
  \centering
  \includegraphics[width=0.9\linewidth,trim={0cm 13.5cm 10cm 0cm},clip]{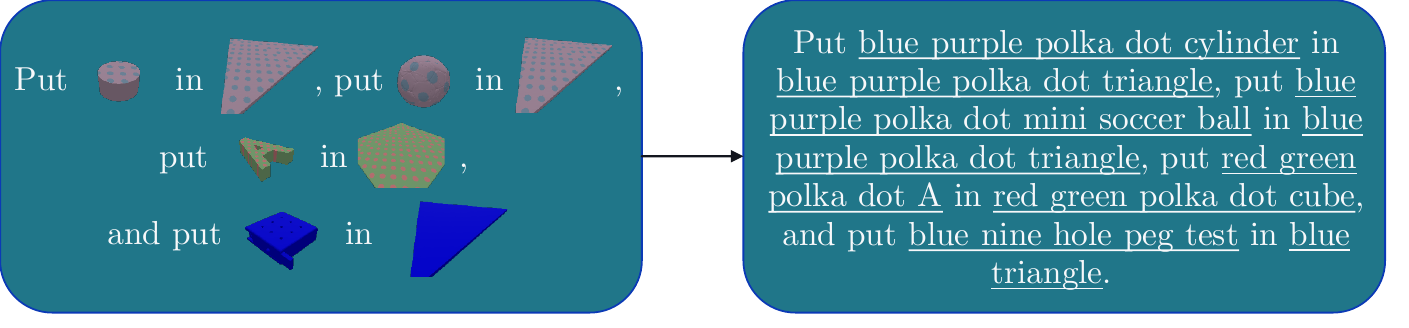}
  \caption{Language-only counterpart of multi-modal prompts achieved by adding simple descriptions of the objects in place of the object image.}
  \label{fig:lang_prompts}
\end{figure}

\begin{table*}[t]
  \centering
  \scriptsize
  \begin{tabularx}{0.85\linewidth}{lccccccccc}
    \toprule
    & \multicolumn{3}{c}{\textbf{L0}} & \multicolumn{3}{c}{\textbf{L1 (zero-shot)}} & \multicolumn{3}{c}{\textbf{L2 (zero-shot)}} \\
    \cmidrule(lr){2-4}
    \cmidrule(lr){5-7}
    \cmidrule(lr){8-10}
    \textbf{Model} & \textbf{Suc.} & \textbf{AR} & \textbf{R/S} & \textbf{Suc.} & \textbf{AR} & \textbf{R/S} & \textbf{Suc.} & \textbf{AR} & \textbf{R/S}\\
    \midrule
    Oracle & 100.0 & 320.00 & 3.06 & 100.0 & 1027.00 & 5.59 & 100.0 & 2583.00 & 9.12  \\
    JAT~\cite{gallouedec2024jack} & 32.5 & 296.17 & 2.68 & 0.0 & 321.60 & 0.98 & 0.0 & 1317.61 & 2.04 \\
    RoboFlamingo~\cite{li2023vision} & 57.5 & 229.30 & 2.98 & 0.0 & 334.55 & 1.01 & 0.0 & 1047.42 & 1.61 \\
    Octo~\cite{team2024octo} & 41.0 & 266.99 &  2.64 & 0.4 & 310.08 & 0.97 & 0.0 & 410.19 & 0.78 \\
    StreamRoboLM (Opt) & 56.0 & 229.50 & 2.85 & 0.0 & 381.78 & 1.12 & 0.0 & 1336.24 & 2.06  \\
    StreamRoboLM (Llama3) & 58.5 & 242.19 & 3.05 & 0.0 & 353.85 & 1.04 & 0.0 & 1181.35 & 1.83 \\
    \bottomrule
  \end{tabularx}
  \caption{Results of language-only prompts baselines.}
  \label{tab:lang_only_results}
\end{table*}

ClevrSkills supports both language-only and multi-modal prompts where multi-modal prompts can be easily converted to language prompts by replacing the ``image'' placeholder with a simple description of the object as shown in Figure~\ref{fig:lang_prompts}. However, some of the task specifications that depend on keysteps can not be described in language only. Therefore, we skip those tasks for these experiments. These tasks include ``Match pose'' and ``Move without hitting'' from L0 and ``Follow order'', ``Follow order and restore'', ``Rearrange'', and ``Rearrange and restore'' from L1. We report the results for all the baselines in Table~\ref{tab:lang_only_results}. 

\rebuttal{We note that all the baselines (except StreamRoboLM) perform better with language only prompts compared to multi-modal prompts. This shows that most SOTA baselines struggle to understand multi-modal prompts. This maybe attributed to suboptimal visual backbones and the fact that most of these baselines are trained on language only task descriptions and therefore are better able to leverage the large-scale pretraining in the language-only scenario.}

\begin{figure}[t!]
    \centering
    \begin{subfigure}{1.0\textwidth}
        \centering
        \includegraphics[width=1.0\linewidth]{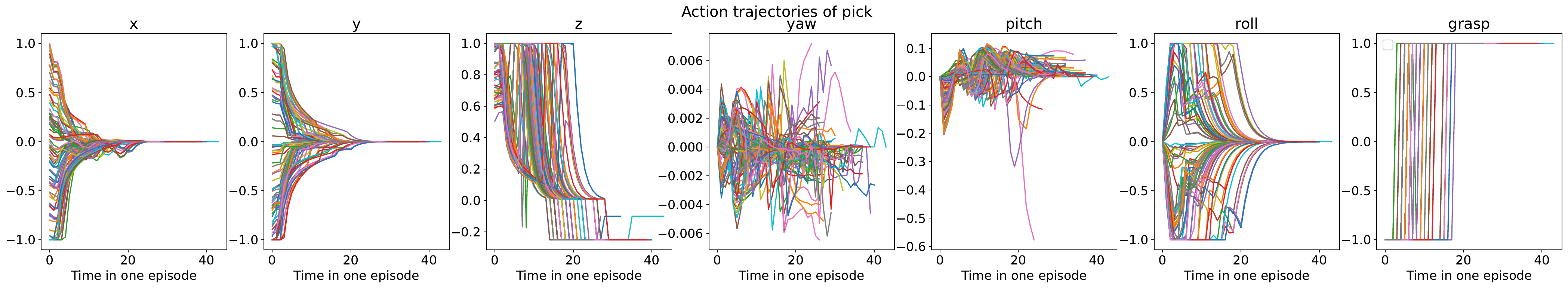}
        \caption{Diversity of action trajectories in ``Pick'' task.}
    \end{subfigure}
    \begin{subfigure}{1.0\textwidth}
        \centering
        \includegraphics[width=1.0\linewidth]{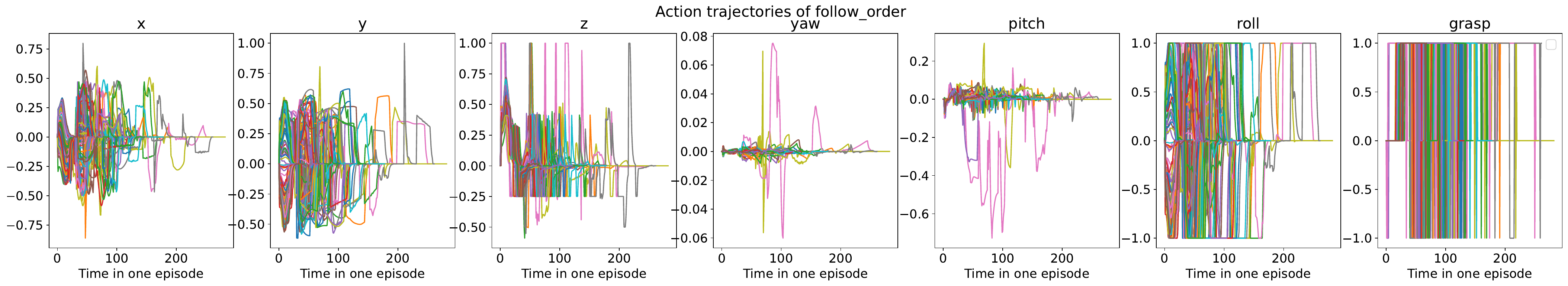}
        \caption{Diversity of action trajectories in ``Follow Order'' task.}
    \end{subfigure}
    \begin{subfigure}{1.0\textwidth}
        \centering
        \includegraphics[width=1.0\linewidth]{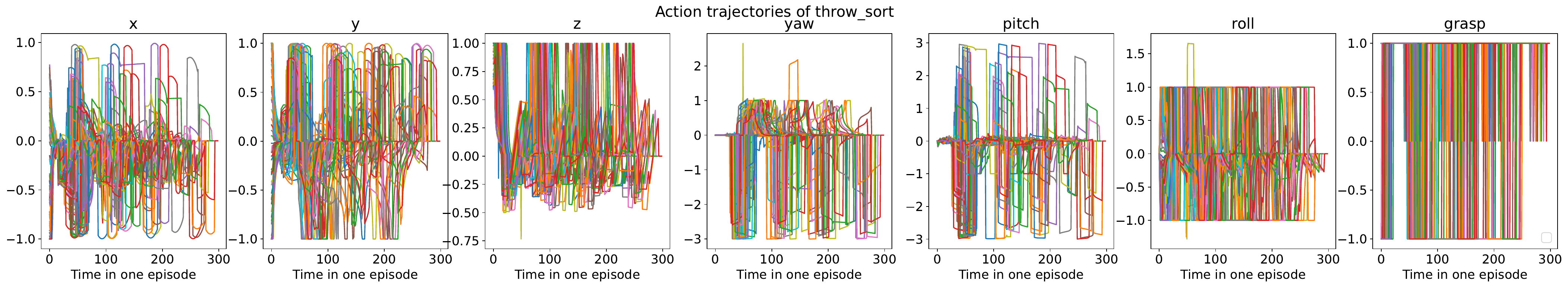}
        \caption{Diversity of action trajectories in ``Throw and Sort" task.}
    \end{subfigure}
    \caption{We plot the action trajectories for 100 randomly sampled episodes each for three different task from the dataset.}
    \label{fig:ep_diversity}
\end{figure}

\rebuttal{
\section{Diversity of ground-truth action trajectories}
We use an optimal motion planner (RTT Connect) to generate near-optimal, canonical trajectories, which nevertheless provide plenty of variance due to stochasticity in the initializations, problem definitions (59 different objects with 61 different textures at random positions) as well as redundancies in the task (eg., target object positions, orientations, etc.).
We plot action trajectories of 100 episodes in one task from each L0, L1 and L2 levels in Figure~\ref{fig:ep_diversity} to show the diversity of the trajectories. As the figures show, just a random sample of 100 episodes (1\% of the data for the particular task) is able to cover a significant portion of the action space in each dimension at each timestamp, showing that the action trajectories from our episodes are highly diverse.}

\section{Datasheet}
\label{sec:datasheet}
We follow \citet{gebru2021datasheets} to provide a datasheet for ClevrSkills below.

\subsection{Motivation}
\textbf{For what purpose was the dataset created?} Was there a specific task in mind? Was there a specific gap that needed to be filled? Please provide a description.

\textit{The dataset was created to carefully study compositional generalization in robotics models i.e. having trained a robotics model on simple manipulation skills, can they generalize to more complicated tasks that require complex compositions of the learned skills.}

\textbf{Who created the dataset (e.g., which team, research group) and on behalf of which entity (e.g., company, institution, organization)?}

\textit{The dataset was created by the authors of the paper on behalf of Qualcomm Technologies Inc.}

\textbf{Who funded the creation of the dataset?} If there is an associated grant, please provide the name of the grantor and the grant name and number.

\textit{N/A}

\textbf{Any other comments?}

\textit{None}

\subsection{Composition}

\textbf{What do the instances that comprise the dataset represent (e.g., documents, photos, people, countries)?}

\textit{The dataset consists of videos of a simulated robot performing and accompanying actions taken by the robot (represented by Nx7 matrix where N is the number of frames and 7 is the dimension of the action vector). The dataset also consists of other annotations including multi-modal prompts for task specification, language annotations for robot actions, object bounding boxes and visibility annotations and frames of key-steps.}

\textbf{How many instances are there in total (of each type, if appropriate)?}

\textit{There are 330k episodes in total spread across 33 different tasks.}

\textbf{Does the dataset contain all possible instances or is it a sample (not necessarily random) of instances from a larger set?}

\textit{The dataset consists of only 10k episodes generated per task. If required, more data can be generated through the ClevrSkills task suite using our oracle policies.}

\textbf{What data does each instance consist of?}

\textit{Each instance consists of a video, as well as corresponding actions (which we consider as labels; see next item).}

\textbf{Is there a label or target associated with each instance?} If so, please provide a description.

\textit{We consider the actions taken by the robot to be the main labels but also include text annotations for the actions, object bounding boxes for all the objects visible to the robot in each frame and images of key-steps.}

\textbf{Is any information missing from individual instances?} If so, please provide a description, explaining why this information is missing (e.g., because it was unavailable). This does not include intentionally removed information, but might include, e.g., redacted text.

\textit{No information is missing from any instance.}

\textbf{Are relationships between individual instances made explicit (e.g., users’ movie ratings, social network links)?} If so, please describe how these relationships are made explicit.

\textit{N/A}

\textbf{Are there recommended data splits (e.g., training, development/validation, testing)?} If so, please provide a description of these splits, explaining the rationale behind them.

\textit{We release the training/validation/testing splits for each of the task in the dataset. The splits only consists of seeds used to generate the episodes therefore, any seeds not used in training or validation set may be used as additional test examples.}

\textbf{Are there any errors, sources of noise, or redundancies in the dataset?} If so, please provide a description. 

\textit{None that the authors are aware of.}

\textbf{Is the dataset self-contained, or does it link to or otherwise rely on external resources (e.g., websites, tweets, other datasets)?} If it links to or relies on external resources, a) are there guarantees that they will exist, and remain constant, over time; b) are there official archival versions of the complete dataset (i.e., including the external resources as they existed at the time the dataset was created); c) are there any restrictions (e.g., licenses, fees) associated with any of the external resources that might apply to a dataset consumer? Please provide descriptions of all external resources and any restrictions associated with them, as well as links or other access points, as appropriate.

\textit{The dataset is used for training models that can then be evaluated in ClevrSkills environment suite. We plan to open-source ClevrSkills environment suite so that any models trained on the data can be evaluated easily.}

\textbf{Does the dataset contain data that might be considered confidential (e.g., data that is protected by legal privilege or by doctor–patient confidentiality, data that includes the content of individuals’ nonpublic communications)?} If so, please provide a description.

\textit{No, the dataset does not contain any confidential data.}

\textbf{Does the dataset contain data that, if viewed directly, might be offensive, insulting, threatening, or might otherwise cause anxiety?} If so, please describe why.

\textit{No, the dataset does not contain any data that might be offensive, insulting, threatening, or might otherwise cause anxiety.}

\textbf{Does the dataset identify any subpopulations (e.g., by age, gender)?} If so, please describe how these subpopulations are identified and provide a description of their respective distributions within the dataset.

\textit{N/A}

\textbf{Is it possible to identify individuals (i.e., one or more natural persons), either directly or indirectly (i.e., in combination with other data) from the dataset?} If so, please describe how.

\textit{No. The dataset does not contain any information about any individual in any form.}

\textbf{Does the dataset contain data that might be considered sensitive in any way (e.g., data that reveals race or ethnic origins, sexual orientations, religious beliefs, political opinions or union memberships, or locations; financial or health data; biometric or genetic data; forms of government identification, such as social security numbers; criminal history)?} If so, please provide a description

\textit{No.}

\textbf{Any other comments?}

\textit{None.}

\subsection{Collection Process}

\subsection{Preprocessing/cleaning/labeling}

\textbf{Was any preprocessing/cleaning/labeling of the data done (e.g., discretization or bucketing, tokenization, part-of-speech tagging, SIFT feature extraction, removal of instances, processing of missing values)?} If so, please provide a description. If not, you may skip the remaining questions in this section.

\textit{After the data generation, any failed trajectories (i.e. the episodes where the oracle policy failed to complete the given task) were discarded. No other preprocessing was done.}

\subsection{Uses}

\textbf{Has the dataset been used for any tasks already?} If so, please provide a description.

\textit{No. This paper is the first instance of the use of the dataset.}

\textbf{Is there a repository that links to any or all papers or systems that use the dataset?} If so, please provide a link or other access point.

\textit{No, such a repository does not exist at this time.}

\textbf{What (other) tasks could the dataset be used for?}

\textit{The dataset may be used for tasks involving robot manipulation}

\textbf{Is there anything about the composition of the dataset or the way it was collected and preprocessed/cleaned/labeled that might impact future uses?} For example, is there anything that a dataset consumer might need to know to avoid uses that could result in unfair treatment of individuals or groups (e.g., stereotyping, quality of service issues) or other risks or harms (e.g., legal risks, financial harms)? If so, please provide a description. Is there anything a dataset consumer could do to mitigate these risks or harms?

\textit{N/A}

\textbf{Are there tasks for which the dataset should not be used?} If so, please provide a description.

\textit{N/A}

\textbf{Any other comments?}

\textit{None.}

\subsection{Distribution}

\textbf{Will the dataset be distributed to third parties outside of the entity (e.g., company, institution, organization) on behalf of which the dataset was created?} If so, please provide a description. 

\textit{Yes. We plan to make our dataset publicly available at \url{https://www.qualcomm.com/developer/software/clevrskills-dataset}.}

\textbf{How will the dataset will be distributed (e.g., tarball on website, API, GitHub)?} Does the dataset have a digital object identifier (DOI)?

\textit{The dataset will be distributed in zip files on our dataset webpage.}

\textbf{When will the dataset be distributed?}

\textit{The dataset is already available on the dataset webpage for the reviewers. The full dataset will be publicly released on the acceptance of this paper.}

\textbf{Will the dataset be distributed under a copyright or other intellectual property (IP) license, and/or under applicable terms of use (ToU)?} If so, please describe this license and/or ToU, and provide a link or other access point to, or otherwise reproduce, any relevant licensing terms or ToU, as well as any fees associated with these restrictions.

\textit{License available on the dataset website. }

\textbf{Have any third parties imposed IP-based or other restrictions on the data associated with the instances?} If so, please describe these restrictions, and provide a link or other access point to, or otherwise reproduce, any relevant licensing terms, as well as any fees associated with these restrictions

\textit{No further restrictions beyond what is mentioned in the license.
}

\textbf{Do any export controls or other regulatory restrictions apply to the dataset or to individual instances?} If so, please describe these restrictions, and provide a link or other access point to, or otherwise reproduce, any supporting documentation.

\textit{N/A}

\textbf{Any other comments?}

\textit{None.}

\subsection{Maintenance}

\textbf{Who will be supporting/hosting/maintaining the dataset?}

\textit{The dataset is hosted and maintained by Qualcomm Technologies Inc.} 

\textbf{How can the owner/curator/manager of the dataset be contacted (e.g., email address)?}

\textit{research.datasets@qti.qualcomm.com}

\textbf{Is there an erratum?} If so, please provide a link or other access point.

\textit{Updates/changes will be specified on the dataset webpage.}

\textbf{Will the dataset be updated (e.g., to correct labeling errors, add new instances, delete instances)?} If so, please describe how often, by whom, and how updates will be communicated to dataset consumers (e.g., mailing list, GitHub)?

\textit{Updates/changes (if any) will be specified on the dataset webpage.}

\textbf{If the dataset relates to people, are there applicable limits on the retention of the data associated with the instances (e.g., were the individuals in question told that their data would be retained for a fixed period of time and then deleted)?} If so, please describe these limits and explain how they will be enforced

\textit{The dataset does not relate to people and therefore we do not foresee a limit on the retention of the data.}

\textbf{Will older versions of the dataset continue to be supported/hosted/maintained?} If so, please describe how. If not, please describe how its obsolescence will be communicated to dataset consumers.

\textit{Yes. All versions should be available at the dataset webpage.}

\textbf{If others want to extend/augment/build on/contribute to the dataset, is there a mechanism for them to do so? If so, please provide a description. Will these contributions be validated/verified}? If so, please describe how. If not, why not? Is there a process for communicating/distributing these contributions to dataset consumers? If so, please provide a description.

\textit{As the dataset is standalone, there is currently no mechanism for extensions. Interested parties are invited to contact the authors about any potential fixes/extensions.}

\textbf{Any other comments?}

\textit{No}

\end{document}